\documentclass[11pt,letterpaper]{article}

\newcommand{\PaperTitle}{Principled Human--AI Collaboration in Agentic Workflows}
\newcommand{\PaperAuthorsPDF}{Young Hyun Cho; Will Wei Sun}
\newcommand{\PaperKeywords}{AI-assisted decision making, black-box systems, statistical calibration}
\newcommand{\PaperAuthorsDisplay}{%
  Young Hyun Cho\textsuperscript{1,\textdagger}\qquad
  Will Wei Sun\textsuperscript{2,*}}
\newcommand{\PaperAffiliations}{%
  \textsuperscript{1}\,Department of Statistics, Harvard University\\
  \textsuperscript{2}\,Daniels School of Business, Purdue University}
\newif\ifShowContactFootnote
\ShowContactFootnotetrue
\newcommand{\YoungEmailNote}{%
  \href{mailto:younghyuncho@fas.harvard.edu}
  {younghyuncho@fas.harvard.edu}}

\newcommand{\WeiEmailNote}{%
  Corresponding author: Will Wei Sun
  (\href{mailto:sun244@purdue.edu}
  {sun244@purdue.edu}).}

\usepackage[letterpaper,margin=1in]{geometry}
\usepackage[T1]{fontenc}
\usepackage[utf8]{inputenc}
\usepackage{lmodern}
\usepackage{microtype}
\usepackage{setspace}
\usepackage{amsmath,amssymb,amsfonts,amsthm,mathtools,mathrsfs}

\usepackage{bm}
\usepackage{bbm}        % permits \mathbbm{1} from older manuscripts

\theoremstyle{plain}
\newtheorem{theorem}{Theorem}
\newtheorem{proposition}{Proposition}
\newtheorem{lemma}{Lemma}

\theoremstyle{definition}

\newtheorem{example}{Example}
\theoremstyle{remark}
\newtheorem{remark}{Remark}

\usepackage{enumitem}
\setlist[enumerate]{leftmargin=*,topsep=0.35em,itemsep=0.18em}
\setlist[itemize]{leftmargin=*,topsep=0.35em,itemsep=0.18em}
\usepackage{graphicx}
\graphicspath{{figures/}{./}}
\usepackage{booktabs,array,tabularx,multirow,longtable}
\usepackage{caption}
\usepackage{subcaption}
\usepackage{algorithm}
\usepackage{algpseudocode}
\makeatletter
\providecommand*{\theHALG@line}{}
\renewcommand*{\theHALG@line}{\theHalgorithm.\arabic{ALG@line}}
\makeatother
\usepackage[most]{tcolorbox}
\tcbuselibrary{breakable}
\usepackage{tikz}
\usetikzlibrary{arrows.meta,positioning,calc}
\usepackage{pgfplots}
\pgfplotsset{compat=1.18}

\usepackage{xcolor}
\definecolor{myblue}{HTML}{6E9FC5}
\definecolor{mybluefill}{HTML}{EEF5FB}
\definecolor{myline}{HTML}{2F3A45}
\definecolor{paperlink}{HTML}{0000FF}

\usepackage{titlesec}
\titleformat{\section}{\normalfont\Large\bfseries}{\thesection}{0.8em}{}
\titleformat{\subsection}{\normalfont\large\bfseries}{\thesubsection}{0.75em}{}
\titleformat{\subsubsection}{\normalfont\normalsize\bfseries}{\thesubsubsection}{0.65em}{}
\titlespacing*{\section}{0pt}{2.0ex plus 0.6ex minus 0.3ex}{0.75ex}
\titlespacing*{\subsection}{0pt}{1.8ex plus 0.4ex minus 0.2ex}{0.5ex}
\titlespacing*{\subsubsection}{0pt}{1.4ex plus 0.3ex minus 0.2ex}{0.35ex}
\usepackage[round,authoryear]{natbib}
\usepackage[unicode,colorlinks=true,linkcolor=paperlink,citecolor=paperlink,urlcolor=paperlink]{hyperref}
\usepackage[nameinlink,noabbrev,capitalise]{cleveref} % optional \cref, \Cref

\hypersetup{
  pdftitle={\PaperTitle},
  pdfauthor={\PaperAuthorsPDF},
  pdfsubject={Research preprint},
  pdfkeywords={\PaperKeywords},
  pdfcreator={pdfLaTeX}
}

\newcommand{\MakePreprintTitle}{%
  \vspace*{-0.38in}% calibrate the title baseline to the reference preprint
  \begin{center}
    {\fontsize{19}{24}\selectfont\normalfont\bfseries
       \PaperTitle\par}
    \vspace{1.20em}
    {\large\normalfont\PaperAuthorsDisplay\par}
    \vspace{2.05em}
    {\normalsize\PaperAffiliations\par}
  \end{center}
  \ifShowContactFootnote
  \begingroup
    \renewcommand{\thefootnote}{\fnsymbol{footnote}}%
    \footnotetext[2]{\YoungEmailNote}%
    \footnotetext[1]{\WeiEmailNote}%
  \endgroup
\fi
  \vspace{0.0em}
}

\newenvironment{preprintabstract}{%
  \begin{center}\begin{minipage}{0.90\textwidth}
  \begin{center}\bfseries Abstract\end{center}
  \small\setstretch{1.05}\ignorespaces
}{%
  \end{minipage}\end{center}\vspace{0.45em}
}

\makeatletter
\newcommand{\SupplementCounter}[1]{%
  \setcounter{#1}{0}%
  \expandafter\renewcommand\csname the#1\endcsname{S\arabic{#1}}%
  \@ifundefined{theH#1}{%
    \expandafter\def\csname theH#1\endcsname{supp.#1.\arabic{#1}}%
  }{%
    \expandafter\renewcommand\csname theH#1\endcsname{supp.#1.\arabic{#1}}%
  }%
}
\makeatother

\newcommand{\BeginSupplementaryMaterial}{%
  \clearpage
  \appendix
  \renewcommand{\thesection}{\Alph{section}}%
  \SupplementCounter{equation}%
  \SupplementCounter{figure}%
  \SupplementCounter{table}%
  \SupplementCounter{algorithm}%
  \SupplementCounter{theorem}%
  \SupplementCounter{proposition}%
  \SupplementCounter{lemma}%
  \SupplementCounter{corollary}%
  \SupplementCounter{observation}%
  \SupplementCounter{definition}%
  \SupplementCounter{assumption}%
  \SupplementCounter{example}%
  \SupplementCounter{remark}%
  \vspace*{-0.30in}% bring the standalone supplement heading higher on page
  \begin{center}
  {\fontsize{19}{23}\selectfont\normalfont\bfseries
     SUPPLEMENTARY MATERIAL of\par}
  \vspace{0em}
  {\fontsize{19}{24}\selectfont\normalfont\bfseries
     ``\PaperTitle''\par}
\end{center}
  \vspace{0.80em}
}

\begin{document}

\MakePreprintTitle

\begin{preprintabstract}
% PASTE ABSTRACT HERE (replace the placeholder sentence).
Organizations adopting AI face the challenge of deciding which outputs can be released without human review while meeting the quality standards of human experts. We develop a statistical framework for delegating tasks between experts and black-box generate–verify workflows, in which AI-generated answers are verified and revised over multiple rounds. Before deployment, human experts evaluate selected outputs, and their judgments are used to calibrate the verifier threshold and number of attempts. Our method selects the policy with the highest estimated automation rate among statistically approved candidates, without modifying the generator or verifier. We establish finite-sample control of the error rate among released outputs, characterize the expert evaluations sufficient to attain a target automation rate, and derive a minimax lower bound establishing rate optimality in the quality margin and required confidence. In a MultiHiertt financial question-answering case study, average held-out automation increases from 2.50\% to 24.17\% as expert evaluations accumulate, while released-output error rates remain around 1.20\%. In contrast, heuristics that release answers after consistently high verifier scores either violate the 5\% error tolerance or automate at most 1.30\% of tasks. Synthetic experiments further examine these theoretical predictions under controlled settings.
\end{preprintabstract}


% ============================================================================
% MAIN PAPER: paste the OR manuscript's main text into the sections below.
% Keep all \label, \ref, \eqref, \citep, \citet commands as appropriate.
% ============================================================================
\section{Introduction}
\label{sec: introduction}

Organizations are increasingly adopting artificial intelligence (AI) to automate recurring tasks traditionally performed by human experts. A recent McKinsey survey reports that 44\% of respondents are scaling AI across their organizations \citep{tinkoff2026state}. If experts must review every AI-generated output, they still need to spend time on every task. On the other hand, using AI-generated outputs without expert review risks accepting answers that do not meet the organization's quality standards. Organizations therefore need a principled way to safely delegate work to AI without requiring human experts to review every output.

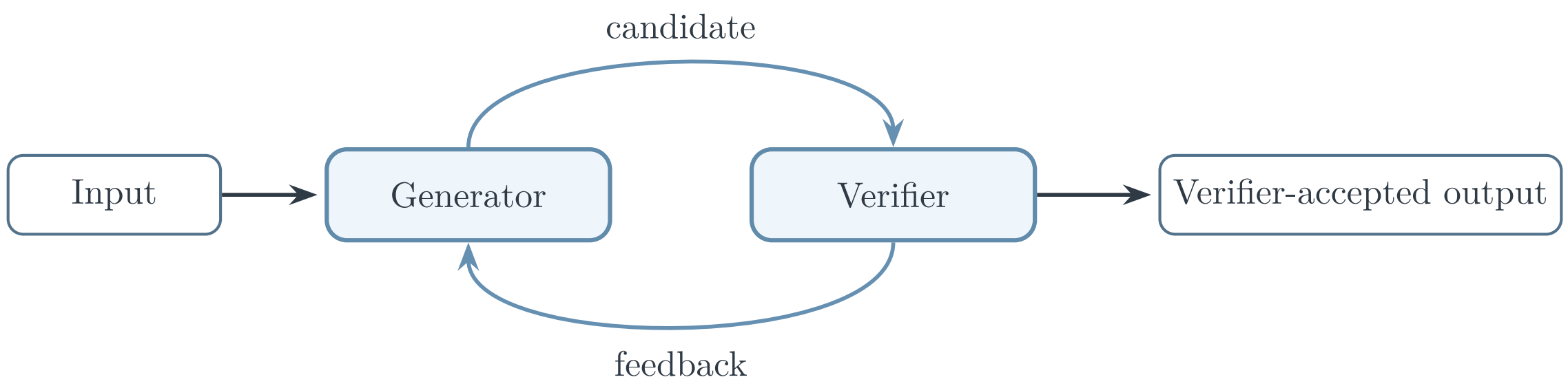
\begin{figure}[t]
\centering
\vspace{0.4em}
\resizebox{0.92\textwidth}{!}{
\begin{tikzpicture}[
    >=Stealth,
    every node/.style={text=myline},
    box/.style={
        draw=myblue!80!myline,
        very thick,
        rounded corners=2mm,
        fill=mybluefill,
        align=center,
        minimum width=2.8cm,
        minimum height=0.9cm,
        font=\small,
        text=myline
    },
    smallbox/.style={
        draw=myblue!55!myline,
        thick,
        rounded corners=1.5mm,
        fill=white,
        align=center,
        minimum width=2.1cm,
        minimum height=0.78cm,
        font=\small,
        text=myline
    },
    arrow/.style={
        -{Stealth[length=2.8mm,width=2mm]},
        line width=1.1pt,
        draw=myline,
        shorten >=2pt
    },
    looparrow/.style={
        ->,
        line width=1.1pt,
        draw=myblue!85!myline
    }
]

% Nodes
\node[smallbox] (input) at (0,0) {Input};
\node[box] (gen) at (3.5,0) {Generator};
\node[box] (eval) at (7.7,0) {Verifier};

% Maintain a fixed gap between Verifier and Output
\node[smallbox, right=1.2cm of eval]
    (out) {Verifier-accepted output};

% Straight arrows
\draw[arrow] (input.east) -- (gen.west);
\draw[arrow] (eval.east) -- (out.west);

% Candidate
\draw[looparrow]
    (gen.north) .. controls +(0,1.1) and +(0,1.1) ..
    node[above=3pt, font=\small, text=myline] {candidate}
    (eval.north);

% Feedback
\draw[looparrow]
    (eval.south) .. controls +(0,-1.1) and +(0,-1.1) ..
    node[below=3pt, font=\small, text=myline] {feedback}
    (gen.south);

\end{tikzpicture}
}
\caption{Generate--verify workflow, adapted from the workflow in \cite{anthropic2024agents}. A generator proposes a candidate, and a verifier assigns a score and provides feedback. The generator may revise the answer in another round if it does not pass verification.}
\label{fig:evaluator_optimizer}
\end{figure}

One common approach to AI automation is a generate--verify workflow, in which AI-generated answers are evaluated and revised over multiple rounds \citep{anthropic2024agents,mckinsey2025_agentic_ai_explained}. Figure~\ref{fig:evaluator_optimizer} illustrates this workflow. A generator proposes an answer, and a verifier assigns a score and provides feedback. If the score is sufficiently high, the answer passes verification; otherwise, the generator can use the feedback to revise it. This process continues until an answer passes verification or the workflow stops. Additional attempts can help AI complete tasks it could not solve in a single round.

Importantly, passing verification does not necessarily mean that an answer meets the organization's quality standards. For example, when a generate--verify workflow is used to answer questions about financial reports, the verifier may accept an answer that uses the wrong reporting year or misinterprets a financial figure. Allowing more attempts may lead to more answers passing verification, but some of these additional answers may still fail to meet the organization's quality standards. This challenge is particularly relevant when organizations rely on off-the-shelf AI systems, where the generator and verifier operate as black boxes whose internal mechanisms are unknown.

A recent industry survey identifies agent quality as a major obstacle to production deployment and finds that human review is widely used to evaluate agent behavior \citep{langchain2026state}. To use AI-generated outputs without reviewing each one, organizations need evidence that the answers they release meet their quality standards. Obtaining this evidence through human expert evaluations requires time and effort before deployment. Organizations therefore need to understand how much human work they can safely delegate to a given AI workflow and how many expert evaluations are needed to justify that delegation.

When organizations use AI to automate work currently performed by human experts, those experts can evaluate AI-generated outputs using the same standards they apply to their own work. Their judgments directly determine whether each evaluated answer meets the organization's quality standard, even when the AI itself remains a black box.

\begin{figure}[t]
\centering
\includegraphics[width=\textwidth]
{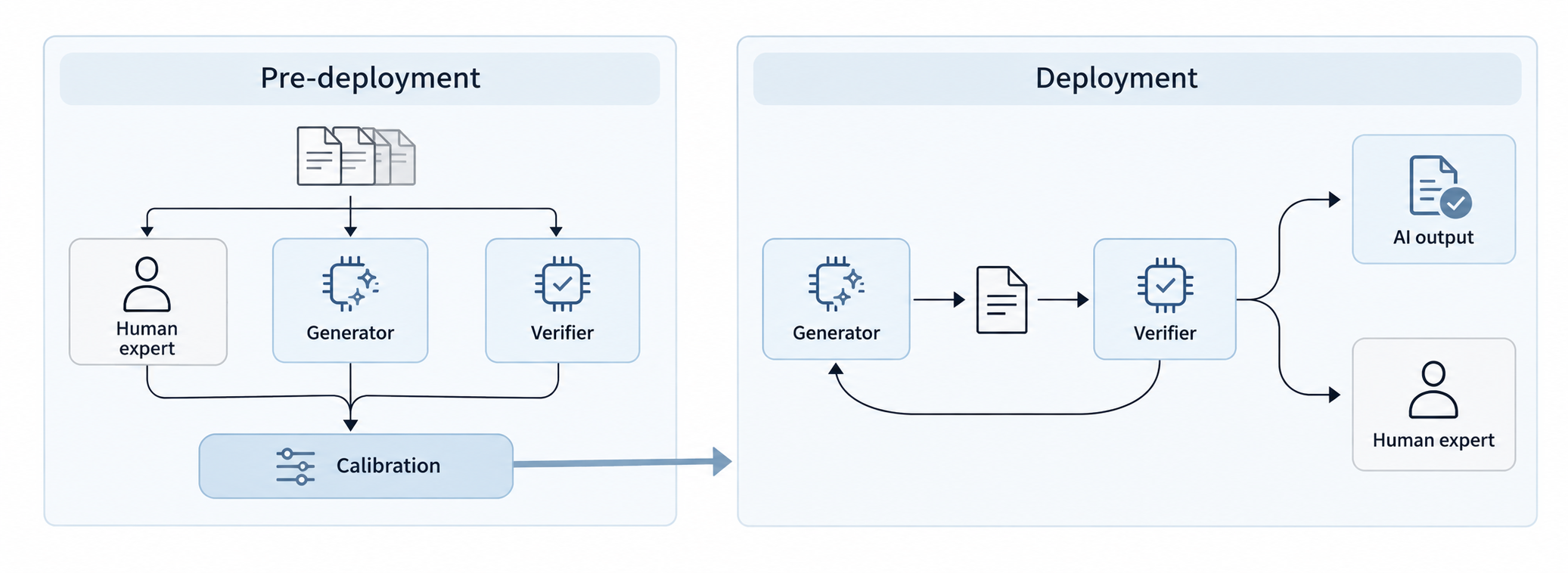}
\caption{Overview of the proposed human--AI collaboration framework. Before deployment, the organization runs the generate--verify workflow on pilot tasks and collects human expert evaluations. These evaluations and the workflow execution records guide the selection of a verifier threshold and round limit. During deployment, the AI releases the first output that passes the selected threshold within the allowed attempts; otherwise, the task is handed to a human expert.}
\label{fig:method_overview}
\end{figure}

In this paper, we develop a framework for safely delegating human work to a given generate--verify workflow using expert evaluations, as illustrated in Figure~\ref{fig:method_overview}. Before deployment, human experts evaluate AI-generated outputs on pilot tasks, and we use their judgments to calibrate a rule for releasing answers without further human review. Our method provides a statistical guarantee, with a prescribed level of confidence, that the error rate among released answers does not exceed the organization's tolerance. Among the rules supported by these evaluations, we select the one estimated to allow AI to handle the largest share of tasks. During deployment, the workflow releases answers according to the selected rule and sends the remaining tasks to human experts.

Moreover, we study safe AI delegation through theory and numerical studies. Our theory quantifies how many expert evaluations suffice for a target automation rate under a quality requirement and shows how additional evaluations recover the highest safe rate among candidate rules. Matching sufficient and minimax lower bounds establish rate optimality in the quality margin and required confidence. In a MultiHiertt financial-report case study \citep{zhao2022multihiertt}, our method increases average held-out automation from 2.50\% to 24.17\% as expert evaluations accumulate, with released-output error rates remaining around 1.20\%. In contrast, heuristics that release answers after consistently high verifier scores either violate the 5\% error tolerance or automate at most 1.30\% of tasks. Synthetic experiments further examine these theoretical predictions under controlled settings.

\subsection{Related Literature and Our Contributions}

Our work builds on research on human--AI collaboration and calibration for AI workflows.

\noindent\textbf{AI Adoption and Human--AI Collaboration.}
Research on AI adoption examines how organizations use AI to automate tasks and augment human work \citep{raisch2021artificial}. Empirical studies show that generative AI can improve worker productivity and task performance \citep{noy2023experimental,brynjolfsson2025generative}. Other work examines changes in human skill requirements, teamwork, and organizational practices \citep{giannitsas2026artificial,devigili2026skill,dell2026cybernetic}. On the other hand, \citet{dell2026navigating} showed that AI assistance improved performance on tasks within its capabilities but reduced performance on tasks beyond them. 

Effective human--AI collaboration depends on when people rely on AI and how tasks are allocated between humans and AI. People may misjudge their abilities relative to AI or struggle to evaluate AI outputs \citep{fugener2022cognitive,lebovitz2022engage}. Other studies examine which decisions to assign to AI or human experts \citep{madras2018predict,mozannar2020consistent,narasimhan2022post,lai2022human,alves2024cost,agarwal2025designing,zhong2026optimal} and when human oversight or verification is needed \citep{fugener2026roles,pmlr-v306-huang26bp,srivastava2026governing}. Related work also studies the efficient use of human feedback for preference alignment, and sequential decision-making \citep{li2021optimal,bhat2024value,zhang2025self,chen2025compo,zhang2025fisher, xiao2025restoring,lu2026contextual}. Our work studies how organizations can use a limited number of expert evaluations to automate as much work as possible with a given AI workflow while controlling the error rate among its released answers.

\noindent\textbf{Agentic AI Workflows and Calibration.}
Research on iterative AI workflows examines generation, feedback, verification, and decisions about computation and stopping \citep{madaan2023self,gou2024critic,snell2024scaling,kalayci2025optimal,dughmi2026adaptive,hegazy2026valid}. Recent work further develops verification-guided revision
and reasoning \citep{liu2026reves,chen2026stepwise}
and confidence-based LLM agents for operations research problems
\citep{li2026coopa}.
Related work studies how to control risks in AI outputs and stopping decisions \citep{lee2024selective,quach2024conformal,jung2025trust,pmlr-v306-wang26hd,lee2026risk}. The approach of \citet{angelopoulos2025learn} is closest to ours in its statistical methodology. They use multiple-testing procedures to approve candidate rules using labeled calibration data. However, such data are not readily available in black-box workflows. We develop a way to collect human expert evaluations that enables valid statistical tests of whether candidate deployment rules meet the organization's quality standards.

%\clearpage
%\subsection{Our Contributions}

Building on these literatures, we make the following contributions:

\noindent\textbf{Problem Formulation.} We formulate organizational AI adoption as a statistical problem of maximizing automation under the quality standards of human experts. In particular, we build on the fact that organizations have experts who can evaluate AI outputs using the same standards they apply to their own work. Rather than assuming these evaluations are already available, we study how many are needed before deployment. This formulation brings a new perspective to the human–AI collaboration literature by linking safe delegation with the expert evaluations needed to support it.

\noindent\textbf{Statistical Methodology.}
We develop a principled method for calibrating black-box generate--verify workflows without modeling the generator or verifier. The key is that human experts can directly judge whether an evaluated answer meets the standards of their work. Their judgments provide the error labels we use for statistical analysis, even though the AI itself remains a black box. We use these evaluations to construct valid tests for candidate deployment rules and select one with a finite-sample guarantee on the error rate among released answers.

\noindent\textbf{Practical Implications.} Our financial question-answering case study shows that heuristics relying on consistently high verifier scores either violate the organization's quality requirement or leave nearly all tasks to human experts. In contrast, our method uses expert evaluations to identify reliable deployment policies, allowing organizations to automate substantially more work while maintaining the required quality standard. Importantly, this improvement requires no retraining or modification of the generator or verifier. Our theoretical results further explain how the expert evaluations needed for safe delegation depend on AI quality.

The remainder of the paper is organized as follows. Section~\ref{sec: problem setup} formulates the human--AI delegation problem and defines the deployment rules that determine when AI-generated answers are released or escalated to human experts. Section~\ref{sec: method} develops our method, describing how expert evaluations are collected and used to select a deployment rule with finite-sample quality protection. Section~\ref{sec: theory} studies how expert evaluations support safe automation, establishes sufficient evaluation requirements, and derives a minimax lower bound on the required human effort. Section~\ref{sec: numerical} presents a financial question-answering case study and synthetic experiments that examine the effects of output quality and automation-rate separation. Section~\ref{sec: conclusion} concludes the paper and discusses directions for future research. The Supplementary Material provides complete proofs and additional experimental details.

\section{Problem Setup}
\label{sec: problem setup}

We consider an organization seeking to automate recurring tasks by adopting off-the-shelf AI. A task \(X\in\mathcal X\) is drawn from \(P_X\). At round \(t\), a generator produces an answer \(C_t\in\mathcal C\), and a verifier assigns a score \(S_t\in[0,1]\). Later answers may depend on earlier answers and scores. Let \(H_0=X\) and \(H_t=(X,C_{1:t},S_{1:t})\) denote the observed history. Write \(C_t\sim P_G(\cdot\mid H_{t-1})\) and \(S_t\sim P_V(\cdot\mid X,C_t)\).

We treat the generator and verifier as black boxes, observing their outputs without assuming parametric models for \(P_G\) or \(P_V\). Both components and their execution rules remain fixed during calibration and deployment. Let \(Q(x,c)\in\{0,1\}\) indicate whether answer \(c\) meets the human expert's standard for task \(x\). An expert evaluation provides the error label \(Y=1-Q(X,C)\), where \(Y=1\) indicates an error. We denote the unknown environment by \(P=(P_X,P_G,P_V,Q)\). The AI does not observe \(Q\) when deciding whether to release an answer.

%\clearpage
\begin{example}[Human-Defined Quality.]
The quality function \(Q(x,c)\) is a key modeling choice that exploits the structure of organizational AI adoption. Expert judgments provide direct binary labels of AI output quality, allowing us to analyze its performance statistically without modeling the AI. The importance of this setting becomes clearer when compared with other AI applications. For mathematical proof agents, verifying a proposed proof may itself require substantial mathematical work, making reliable labels difficult to obtain. On the other hand, coding agents can be evaluated through unit tests, but passing many tests does not establish correctness in untested cases. In our setting, expert evaluations directly measure the quality we seek to control.    
\end{example}

We control how many attempts the AI may make and which verifier scores count as passing. Fix a maximum number of rounds \(T\) and a finite threshold set
\(\mathcal S=\{s_1,\ldots,s_L\}\subset[0,1]\). For each \(s\in\mathcal S\), let \(\tau_s=\inf\{t\in\{1,\ldots,T\}:S_t\ge s\}\), with \(\tau_s=\infty\) if no answer passes. Given a round limit \(m\in\{1,\ldots,T\}\), the rule \(\pi^{(s,m)}\) releases \(C_{\tau_s}\) when \(\tau_s\le m\) and otherwise sends the task to a human expert. Thus, the threshold \(s\) determines which answers pass verification, while \(m\) determines how many attempts the AI may make before escalation.

For an environment \(P\), define the automation rate \(r_P(\pi)\) as the probability that rule \(\pi\) releases an answer on a fresh task, and \(b_P(\pi)\) as the probability that it releases an answer with \(Q=0\). For candidate rules, write \(r_m(s)=r_P(\pi^{(s,m)})\) and \(b_m(s)=b_P(\pi^{(s,m)})\). The error rate among released answers is \(e_m(s)=b_m(s)/r_m(s)\) when \(r_m(s)>0\). We also include the escalation-only rule \(\pi^{\mathrm{esc}}\), which sends every task to an expert and satisfies \(r_P(\pi^{\mathrm{esc}})=b_P(\pi^{\mathrm{esc}})=0\). The next proposition shows how the threshold and round limit affect automation and output quality.

\begin{proposition}[Release ordering and output quality]
\label{prop: round budgets}
For any environment \(P\), thresholds \(s,s'\in\mathcal S\) with \(s\le s'\), and round limits \(1\le m\le m'\le T\), we have
\[
r_m(s')\le r_m(s)\le r_{m'}(s).
\]
In contrast, the error rate \(e_m(s)\), when defined, need not be monotone in either \(s\) or \(m\).
\end{proposition}

Proposition~\ref{prop: round budgets} shows that lowering the verifier threshold or allowing more rounds can increase the automation rate, but neither guarantees better output quality. Additional rounds introduce new releases whose quality may differ from earlier ones, while changing the threshold can also change which answer is released. Thus, automation rates and verifier scores alone cannot determine whether a deployment rule meets the organization's quality standards.

We therefore seek to maximize the automation rate while keeping the error rate among released answers below the organization's tolerance. For \(\alpha\in(0,1)\), the population optimization problem is
\begin{equation}
\label{eq: population objective}
\begin{aligned}
\underset{s\in\mathcal S,\;1\le m\le T}{\operatorname{maximize}}
&\quad r_m(s)\\
\text{subject to} &\quad b_m(s)\le\alpha r_m(s).
\end{aligned}
\end{equation}

For \(r_m(s)>0\), the constraint is equivalent to \(e_m(s)\le\alpha\), and it also holds for zero-release rules. If no candidate is selected, escalation-only remains available. The optimization thus describes the maximum safe automation achievable if the population quantities were known.

%In practice, these quantities are unknown, and obtaining expert evaluations requires effort. We develop a method that uses expert evaluations from pilot tasks to select a deployment rule that protects output quality while maximizing automation.

\section{Proposed Method}
\label{sec: method}

We use expert evaluations of pilot answers to select a deployment rule that meets the organization's quality standard. Let \(n\) be the number of expert evaluations and \(\widehat\pi_n\) the selected rule. For \(\delta\in(0,1)\), we seek the finite-sample guarantee
\begin{equation}
\label{eq: quality requirement}
\mathbb P_P\!\left(
b_P(\widehat\pi_n)\le\alpha r_P(\widehat\pi_n)
\right)\ge1-\delta.
\end{equation}
Here, \(\alpha\) limits the error rate among AI-released answers, while \(\delta\) controls the probability of selecting a rule that violates this limit over the randomness in calibration. We use expert evaluations to test candidate rules simultaneously and select the approved rule with the highest estimated automation rate. Algorithm~\ref{alg: human calibrated workflow} summarizes the complete procedure.

\subsection{Human Evaluations and First-Pass Records}
\label{sec: first pass records}

We evaluate the answers that candidate rules would actually release. For each independent pilot task \(X_\ell\sim P_X\), we assign a threshold index \(J_\ell\) uniformly at random from \(\{1,\ldots,L\}\), independently of the task and previous executions. We run the workflow at threshold \(s_{J_\ell}\) until its first verifier pass or round \(T\). Let \(\tau_\ell\) denote the first-pass round, with \(\tau_\ell=\infty\) if no answer passes. When a pass occurs, an expert evaluates \(C_{\ell,\tau_\ell}\) and provides the error label \(Y_\ell=1-Q(X_\ell,C_{\ell,\tau_\ell})\). We retain \((J_\ell,\tau_\ell)\) even when no answer passes.

We stop collecting pilot tasks after exactly \(n\) expert evaluations, whether the answers are correct or incorrect. Let \(N\) be the total number of pilot tasks executed. We assume \(\sum_{j=1}^L r_T(s_j)>0\), so each execution has a positive probability of producing a verifier pass and \(N\) is finite almost surely.

If a task assigned to \(s_j\) first passes at round \(t\), all rules with \(m\ge t\) release the same answer, while those with \(m<t\) escalate. Thus, a single expert evaluation can be reused across all round limits \(m\ge t\) at the same threshold. Across different thresholds, however, the first passing answer may differ, so an evaluation at one threshold cannot generally be reused at another. Random threshold assignment allows us to collect evaluations for each threshold without evaluating every task under every candidate rule.

After collecting \(n\) expert evaluations from \(N\) pilot tasks, we use the records to assess each candidate deployment rule. Let \(N_j\) count the pilot tasks assigned to threshold \(s_j\), \(R_{j,m}\) those that first pass by round \(m\), and \(B_{j,m}\) the errors among their evaluated answers. Thus, \(\sum_j N_j=N\) and \(\sum_j R_{j,T}=n\). For \(N_j>0\), we estimate the automation rate by \(\widehat r_m(s_j)=R_{j,m}/N_j\). Tasks without a verifier pass remain in the denominator, while only released answers contribute to the quality assessment.

To assess a deployment rule's error rate, we need evaluations of the answers that the rule would actually release. Our first-pass records provide exactly these evaluations. Although the number \(N\) of pilot executions is random, the \(n\) evaluated records are independent draws from the distribution conditional on a verifier pass. For each candidate \((s_j,m)\), the relevant labels therefore follow the rule's released-output distribution, giving, whenever \(r_m(s_j)>0\),
\[
B_{j,m}\mid R_{j,m}\sim\operatorname{Bin}\!\left(R_{j,m},e_m(s_j)\right).
\]

This exact law requires independence across pilot tasks, not across rounds within a task, and thus allows the generator to use earlier answers and verifier feedback. The Supplementary Material provides the proof, including the effect of random stopping.

\subsection{Simultaneous Quality Approval}
\label{sec: quality approval}

We now use the pilot evaluations to assess which deployment rules meet the organization's quality standard. For each candidate \((s_j,m)\) with \(r_m(s_j)>0\), we test the null hypothesis \(H_{0,j,m}:e_m(s_j)\ge\alpha\). When \(R_{j,m}>0\), the conditional binomial law gives the exact lower-tail p-value
\begin{equation}
\label{eq: direct binomial pvalue}
p_{j,m}=\mathbb P\!\left(
\operatorname{Bin}(R_{j,m},\alpha)\le B_{j,m}\right).
\end{equation}
Fewer observed errors provide evidence that the true error rate is below \(\alpha\). When \(R_{j,m}=0\), we set \(p_{j,m}=1\), so a candidate without evaluated releases cannot be approved. Under the null, the p-value is super-uniform, satisfying \(\mathbb P_P(p_{j,m}\le u)\le u\) for all \(u\in[0,1]\). Although \(e_m(s_j)=\alpha\) meets the population quality constraint, it belongs to the null, so approval requires evidence that the error rate is strictly below \(\alpha\).

However, testing each candidate separately does not protect the final deployment decision. With \(LT\) candidates, even valid individual tests can lead to false approvals when considered together \citep{ramdas2025hypothesis}. We therefore control the familywise error rate (FWER), the probability of approving at least one candidate whose error rate is at least \(\alpha\). Since candidate rules may share human evaluations, their p-values can be dependent. This simultaneous control allows us to select among the approved rules using the same pilot data.

We use weighted Holm to control FWER without assuming independence among the tests \citep{holm1979simple}. Before pilot collection, we assign each candidate a positive weight \(\omega_{j,m}\), with \(\sum_{j,m}\omega_{j,m}=1\). We order candidates by increasing \(p_{j,m}/\omega_{j,m}\), breaking ties lexicographically in \((j,m)\). Let \(i_k=(j_k,m_k)\) denote the candidate at position \(k\), and \(W_k=\sum_{h=k}^{LT}\omega_{i_h}\) the remaining weight. Starting with the first candidate, we approve \(i_k\) when
\(
p_{i_k}\le \frac{\delta\omega_{i_k}}{W_k}.
\) At the first failure, we stop and leave that candidate and all subsequent candidates unapproved. The procedure controls FWER at level \(\delta\) using valid p-values and weights fixed before calibration.

At each threshold, the full-horizon rule has the highest automation rate, although it may not meet the quality standard. We therefore reserve a fixed share of the testing weight for full-horizon candidates to support their approval when safe. For \(T>1\), we set
\begin{equation}
\label{eq: weighted testing weights}
\omega_{j,m}
=
\begin{cases}
\dfrac{1}{3L}, & m=T,\\[4pt]
\dfrac{2}{3L(T-1)}, & m<T.
\end{cases}
\end{equation}
For \(T=1\), we set \(\omega_{j,1}=1/L\). Under weighted Holm, \(p_{j,m}\le\delta\omega_{j,m}\) is sufficient for approval because the remaining testing weight never exceeds one. Thus, for \(T>1\), a full-horizon candidate is approved whenever \(p_{j,T}\le\delta/(3L)\), regardless of how many shorter round limits are considered.

\subsection{Deployment and Quality Guarantee}
\label{sec: calibration deployment}

Let \(\widehat{\mathcal C}_n\) denote the set of threshold--round-limit pairs approved by weighted Holm. If \(\widehat{\mathcal C}_n\ne\varnothing\), we choose
\begin{equation}
\label{eq: selected round budget}
(\widehat s,\widehat m)
\in
\operatorname*{arg\,max}_{(s_j,m)\in\widehat{\mathcal C}_n}
\widehat r_m(s_j)
\end{equation}
and set \(\widehat\pi_n=\pi^{(\widehat s,\widehat m)}\). Ties favor the larger round limit and then the smaller threshold index. If no candidate is approved, we select \(\widehat\pi_n=\pi^{\mathrm{esc}}\).

By release ordering, \(\widehat r_m(s_j)\) is nondecreasing in \(m\). Thus, we need only compare the largest approved round limit at each threshold. In particular, when \(L=1\), we select the largest approved round limit, if any. The selected rule remains fixed during deployment.

We emphasize that human escalation serves a practical purpose beyond maintaining the quality guarantee. It can be particularly valuable in high-stakes applications, where incorrect AI-generated answers may have serious consequences. For example, inaccurate automated responses in financial services can mislead customers or cause financial harm \citep{cfpb2023chatbots}, while clinical decisions may depend on patient information missing from AI-generated responses \citep{ha2024artificial}.

\begin{algorithm}[h!]
\caption{Human-Calibrated Generate--Verify Workflow}
\label{alg: human calibrated workflow}

\algtext*{EndIf}
\algtext*{EndFor}
\algtext*{EndWhile}

\begin{algorithmic}[1]
    \State \textbf{Input:} Fresh tasks, generator, verifier, human evaluator, and parameters \(\mathcal S,T,n,\alpha,\delta\)

    \Statex \textbf{Pilot collection.}
    \State Fix the testing weights \(\omega_{j,m}\) as specified in Section~\ref{sec: quality approval}
    \State Set \(N\gets0\) and \(i\gets0\)
    \While{\(i<n\)}
        \State Set \(N\gets N+1\)
        \State Draw \(X_N\sim P_X\) and \(J_N\) uniformly from \(\{1,\ldots,L\}\)
        \State Run the workflow at \(s_{J_N}\) until its first pass or round \(T\)
        \State Record \(\tau_N=\infty\) if no pass occurs
        \If{\(\tau_N\le T\)}
            \State Obtain \(Y_N=1-Q(X_N,C_{N,\tau_N})\) and set \(i\gets i+1\)
        \EndIf
        \State Retain \((J_N,\tau_N)\), together with \(Y_N\) when a pass occurs
    \EndWhile

    \Statex \textbf{Simultaneous quality approval.}
    \State Compute \(N_j,R_{j,m},B_{j,m}\) and \(p_{j,m}\) for all candidates by \eqref{eq: direct binomial pvalue}, setting \(p_{j,m}=1\) when \(R_{j,m}=0\)
    \State Order candidates as \(i_1,\ldots,i_{LT}\) by increasing \(p_{j,m}/\omega_{j,m}\), breaking ties lexicographically
    \State Set \(W\gets1\) and \(\widehat{\mathcal C}_n\gets\varnothing\)
    \For{\(k=1,\ldots,LT\)}
        \If{\(p_{i_k}>\delta\omega_{i_k}/W\)}
            \State \textbf{break}
        \EndIf
        \State Add \((s_{j_k},m_k)\) to \(\widehat{\mathcal C}_n\)
        \State Set \(W\gets W-\omega_{i_k}\)
    \EndFor

    \Statex \textbf{Deployment rule selection.}
    \If{\(\widehat{\mathcal C}_n\ne\varnothing\)}
        \State Choose \((\widehat s,\widehat m)\) maximizing \(R_{j,m}/N_j\) over \(\widehat{\mathcal C}_n\)
        \State Break ties by larger \(m\), then smaller \(j\)
        \State Set \(\widehat\pi_n\gets\pi^{(\widehat s,\widehat m)}\)
    \Else
        \State Set \(\widehat\pi_n\gets\pi^{\mathrm{esc}}\)
    \EndIf
\end{algorithmic}
\end{algorithm}

%\clearpage
We now establish that the selected rule meets the finite-sample quality requirement, even though the same pilot data are used to approve candidates and select the final rule.

\begin{theorem}[Quality Protection]
\label{thm: release quality}
Fix \(\alpha,\delta\in(0,1)\), a finite threshold set \(\mathcal S\subset[0,1]\), and a positive integer \(T\). For every environment \(P\) satisfying \(\sum_{j=1}^L r_T(s_j)>0\) and every fixed positive integer \(n\), Algorithm~\ref{alg: human calibrated workflow} satisfies
\begin{equation}
\label{eq: release quality guarantee}
\mathbb P_P\!\left( b_P(\widehat\pi_n)\le \alpha r_P(\widehat\pi_n) \right)\ge1-\delta.
\end{equation}
\end{theorem}

Theorem~\ref{thm: release quality} guarantees that the selected deployment rule meets the organization's quality standard with probability at least \(1-\delta\). This guarantee holds even when the same pilot data are used to approve candidate rules and select the one with the highest estimated automation rate. It is noteworthy that this guarantee requires no parametric model for the generator or verifier and allows later answers to depend on earlier feedback.

We highlight that Theorem~\ref{thm: release quality} complements research on human--AI task allocation with a finite-sample guarantee for safe delegation. \citet{fugener2026roles} study task allocation to improve organizational performance, while \citet{ahsen2026will} develop policies that account for patient safety in healthcare. Theorem~\ref{thm: release quality} shows how organizations can establish, before deployment, that their selected delegation rule meets human experts' quality standards. This guarantee uses a finite number of expert evaluations without modeling the AI or assuming that verifier scores reliably measure output quality.

\begin{remark}[Planning Human Evaluation and AI Execution]
\label{rem: organizational resources}
A practical advantage of our framework is that organizations can adopt existing AI workflows without post-training the generator or verifier for their own tasks. Such training may require specialized AI expertise and resources, making it difficult for organizations without these capabilities. Instead, the main additional human effort comes from existing experts evaluating AI-generated answers using their usual work standards. Organizations can fix the number \(n\) of expert evaluations, while the total number \(N\) of pilot executions needed to collect them depends on how often answers pass verification. These upfront evaluations can help organizations safely delegate more work to AI, reducing the tasks left to human experts after deployment without substantially changing existing work processes.
\end{remark}

Overall, our method uses expert evaluations to identify safe deployment rules and selects the one with the highest estimated automation rate. This leaves practical questions about how many evaluations are sufficient to achieve a desired automation rate, how AI quality affects this effort, and what organizations can gain from additional evaluations.% In the next section, we address these questions and study the fundamental limits on the expert evaluations required for safe AI delegation.

\section{Theoretical Results}
\label{sec: theory}

In Section~\ref{sec: method}, we established a finite-sample guarantee that our deployment rule meets the organization's quality requirement for any fixed number \(n\) of expert evaluations. An important feature of our framework is that tasks can be escalated to human experts. At the extreme, escalating every task satisfies the quality requirement without releasing any AI-generated answers. This fallback serves as a safeguard, but relying on it alone offers no automation and is therefore less compelling to organizations seeking to automate work.

We thus study how expert evaluations support safe automation. We first show, with sufficient evaluations, our method recovers the highest safe automation rate attainable by the given workflow among the candidate rules. We then quantify how many expert evaluations are sufficient to achieve a desired automation rate while preserving the quality guarantee. Finally, we establish a lower bound showing that the dependence of this evaluation effort on the quality margin and confidence levels cannot be improved in order, even by more flexible evaluation and deployment methods.

Throughout this section, we use \(O(\cdot)\), \(\Omega(\cdot)\), and \(\Theta(\cdot)\) to denote upper, lower, and matching-order bounds up to constant factors, respectively. Each result specifies which problem parameters these constants may depend on.

\subsection{Recovery of Optimal Deployment}

For an environment \(P\), let \(r_P^\star\) denote the highest safe automation rate attainable by the given workflow among our threshold--round-limit rules, with \(r_P^\star=0\) if no candidate satisfies the population quality constraint. Since the true error and automation rates are unknown, the organization cannot directly identify an optimal rule. We examine whether our method can nevertheless recover this population optimum as expert evaluations accumulate.

\begin{theorem}[Recovery of Optimal Deployment]
\label{thm: policy selection consistency}
Fix \(P,\mathcal S,T\) and \(\alpha,\delta\in(0,1)\), where \(\mathcal S\) is finite and \(T\) is a positive integer. Suppose at least one candidate \((s_j,m)\) satisfies \(r_m(s_j)=r_P^\star>0\) and \(e_m(s_j)<\alpha\). Then, as \(n\to\infty\),
\begin{equation}
\label{eq: policy selection consistency}
\mathbb P_P\!\left(b_P(\widehat\pi_n)\le\alpha r_P(\widehat\pi_n),\ r_P(\widehat\pi_n)=r_P^\star\right)\longrightarrow1.
\end{equation}
Moreover, \(\mathbb E_P[r_P(\widehat\pi_n)]\to r_P^\star\). If the optimal feasible candidate is unique, the probability of selecting that pair tends to one.
\end{theorem}

Theorem~\ref{thm: policy selection consistency} shows that sufficient expert evaluations allow an organization to achieve the highest safe automation rate among its candidate rules without changing the generator or verifier. An organization may delegate relatively little work to AI for two different reasons. The existing workflow may lack a rule that can safely automate a large share of tasks, or such a rule may exist but the organization lacks sufficient expert evaluations to approve it. In the first case, additional evaluations cannot improve the workflow's capabilities. In the second, they can help the organization make better use of the safe automation already available. Thus, human effort invested before deployment can help organizations automate more work and reduce the tasks left to human experts afterward.

The strict quality condition in the theorem is important because a candidate with error rate exactly \(\alpha\) satisfies the population quality constraint but remains under the testing null. Its probability of approval therefore cannot exceed \(\delta\), regardless of the number of evaluations. In contrast, when at least one optimal candidate is strictly safe, it is eventually approved, while unsafe candidates are excluded and estimated automation rates converge to their population values. Since the candidate set is finite, these properties allow our method to recover the optimal safe automation rate.

\subsection{Expert Effort for Targeted Automation}
\label{sec: sufficient evaluations}

While Theorem~\ref{thm: policy selection consistency} shows that our method can eventually recover the highest safe automation rate among the considered deployment rules, organizations do not necessarily need to achieve this optimum. A lower level of automation may already meet their needs, and certifying a rule that reaches this level may require fewer expert evaluations than certifying an optimal rule. We therefore study how many expert evaluations are sufficient to achieve a desired automation rate while preserving output quality. Let \(r_{\mathrm{tar}}\in(0,1]\) denote this target. For \(\eta\in(0,1)\), we seek \(\mathbb P_P(r_P(\widehat\pi_n)\ge r_{\mathrm{tar}})\ge1-\eta\). The target \(r_{\mathrm{tar}}\) is not an input to Algorithm~\ref{alg: human calibrated workflow}; it is used only to assess the automation performance of the selected rule.

Our method uses expert evaluations to assess output quality and AI execution records to estimate automation rates. For each candidate \((s_j,m)\), the counts \(R_{j,m}\) and \(B_{j,m}\) provide information about its error rate, while \(N_j\) and \(R_{j,m}\) determine its estimated automation rate. Since pilot collection stops after \(n\) expert evaluations, both types of information accumulate together. Although our method does not require knowledge of the underlying population rates, we introduce two margins based on these unknown quantities to characterize the sufficient evaluation budget.

The first margin concerns the quality of AI-released answers. Even when a rule can automate the desired share of tasks, its safety must be established through expert evaluations. A rule whose error rate is well below the organization's tolerance is easier to certify than one whose error rate barely meets it. We denote by \(\Delta_e>0\) a lower bound on this quality margin, so a target-achieving candidate satisfies \(e_m(s_j)\le\alpha-\Delta_e\).

The second margin concerns the comparison of automation rates. Since our method selects the approved candidate with the highest estimated automation rate, sampling variation may cause a below-target candidate to be selected over a target-achieving candidate when their automation rates are close. More AI execution records are needed to distinguish these candidates reliably. We denote by \(\Delta_r>0\) a lower bound on the automation-rate difference between the target-achieving candidate and below-target candidates at other thresholds. Within the same threshold, release ordering and our tie rule already resolve these comparisons.

The next theorem quantifies how these two margins affect the expert evaluation budget.

\begin{theorem}[Expert Evaluations for Joint Deployment]
\label{thm: sufficient evaluations}
Fix \(r_{\mathrm{tar}}\in(0,1]\), \(0<\Delta_e<\alpha<1\), \(\Delta_r\in(0,1]\), and \(\delta,\eta\in(0,1/4]\). Suppose an environment \(P\) admits a candidate \((s_j,m)\) with \(r_m(s_j)\ge r_{\mathrm{tar}}\) and \(e_m(s_j)\le\alpha-\Delta_e\), and that \(r_m(s_j)-r_k(s_h)\ge\Delta_r\) whenever \(h\ne j\), \(1\le k\le T\), and \(r_k(s_h)<r_{\mathrm{tar}}\). Then there exists \(\bar n=O\!\left(\frac{L\log(LT/(\delta\eta))}{r_{\mathrm{tar}}\Delta_e^2}+\frac{L\log(L/\eta)}{\Delta_r^2}\right)\) such that, for every integer \(n\ge\bar n\), Algorithm~\ref{alg: human calibrated workflow} satisfies \(\mathbb P_P(r_P(\widehat\pi_n)\ge r_{\mathrm{tar}})\ge1-\eta\). Together with Theorem~\ref{thm: release quality}, this gives
\begin{equation}
\label{eq: joint deployment guarantee}
\mathbb P_P\!\left(b_P(\widehat\pi_n)\le\alpha r_P(\widehat\pi_n),\ r_P(\widehat\pi_n)\ge r_{\mathrm{tar}}\right)\ge1-\delta-\eta.
\end{equation}
\end{theorem}

Theorem~\ref{thm: sufficient evaluations} shows that achieving a desired automation rate does not require identifying or certifying the optimal deployment rule. Instead, it is sufficient to approve a candidate that meets the target and ensure that errors in estimated automation rates do not cause a below-target candidate to be selected. This distinction matters because a target-achieving rule may be substantially easier to certify than an optimal rule. The theorem establishes how many evaluations are sufficient to achieve this target while maintaining the organization's quality standard.

The first term in the bound for \(\bar n\) accounts for the expert evaluations needed to certify a target-achieving candidate. Although the organization collects \(n\) evaluations, only those associated with answers that a particular candidate would release inform its error rate. For candidate \((s_j,m)\), their expected number is \(\mathbb E_P[R_{j,m}]=nr_m(s_j)/\sum_{h=1}^L r_T(s_h)\), which is at least \(nr_{\mathrm{tar}}/L\) when the candidate meets the target. A smaller quality margin \(\Delta_e\) makes certification harder, explaining its inverse-square dependence in the bound. Increasing the round limit can provide more evaluated releases, but if the additional answers are less accurate, the resulting quality margin may shrink. Thus, higher automation does not necessarily make quality certification easier.

The second term in the bound for \(\bar n\) addresses the difficulty of comparing automation rates. When \(\Delta_r\) is small, more AI execution records are needed to distinguish a target-achieving candidate from below-target candidates. These comparisons do not require human error labels, but our pilot stops after \(n\) expert evaluations, so increasing \(n\) also provides more execution records. When \(L=1\), there are no cross-threshold comparisons. Since automation rates are nondecreasing in the round limit and our tie rule favors larger limits, an approved target-achieving candidate cannot lose to a below-target candidate at the same threshold. The second term and its separation condition can thus be omitted. More generally, adding thresholds expands the available deployment rules but also spreads pilot observations across more candidates, as reflected in the sufficient evaluation bound.

\begin{remark}[AI Quality and Expert Evaluation Effort]
\label{rem: ai quality and expert effort}
The quality margin \(\Delta_e\) deserves particular attention because it captures the statistical difficulty that directly requires human expert evaluations. Automation rates can be estimated from AI execution records, but establishing whether released answers meet the organization's quality standard requires expert judgments. To isolate this difficulty, we hold the target automation rate, candidate set, confidence levels, and a positive automation-rate separation \(\Delta_r\) fixed, and examine how the sufficient evaluation budget depends on \(\Delta_e\). This comparison is meaningful because verifier pass rates and human-assessed output quality need not change together. Under these fixed conditions, the sufficient evaluation budget has \(O(\Delta_e^{-2})\) dependence as \(\Delta_e\to0\), with quality certification eventually dominating the bound. This reveals an additional benefit of improving AI output quality. Beyond potentially allowing organizations to automate more tasks, better answers can reduce the expert effort sufficient to certify a desired level of safe automation before deployment. As we show in the next subsection, the inverse-square dependence on the quality margin also appears in the minimax lower bound, establishing that this difficulty is not specific to our calibration procedure.
\end{remark}

\subsection{Fundamental Limits on Expert Evaluation}
\label{sec: evaluation lower bound}

Theorem~\ref{thm: sufficient evaluations} provides a sufficient evaluation budget for achieving a desired automation rate while protecting output quality. We now ask whether this requirement reflects a limitation of our method or a fundamental difficulty of safe AI delegation. Our method assigns thresholds uniformly, evaluates first-passing answers, and selects from a fixed set of deployment rules. More flexible procedures could adaptively choose which answers to evaluate and make deployment decisions using the full workflow history. We investigate whether such flexibility can fundamentally reduce the expert evaluations needed to guarantee both quality and automation.

To make this comparison, we consider procedures that can execute the AI workflow without restriction when human evaluations are not requested. They may select answers for expert evaluation adaptively and use any measurable, history-dependent rule to release, continue, or escalate at each round. During deployment, no new human-quality information is available. Each procedure must terminate almost surely and use at most \(n\) expert evaluations. We require a single procedure to satisfy the quality guarantee in Theorem~\ref{thm: release quality} for every admissible environment and the target-automation guarantee in Theorem~\ref{thm: sufficient evaluations} whenever its conditions hold. Let \(n^\star\) be the smallest evaluation budget for which a procedure satisfies both requirements. Since our method belongs to this broader class, the preceding sufficient bound also provides an upper bound on \(n^\star\).

\begin{theorem}[Fundamental Lower Bound on Expert Evaluation]
\label{thm: lower bound}
Assume that \(\mathcal X\) is an infinite standard Borel space and that calibration and deployment rules are measurable. Under the preceding information model, fix \(\alpha\in(0,1)\), a finite threshold set \(\mathcal S\subset[0,1]\), a positive integer \(T\), \(r_{\mathrm{tar}}\in(0,1]\), and \(\Delta_r\in(0,1]\). For \(\Delta_e\in(0,\alpha/2]\) and \(\delta,\eta\in(0,1/4]\),
\begin{equation}
\label{eq: evaluation rate optimality}
n^\star=\Omega\!\left(\frac{\log(1/\delta)+\log(1/\eta)}{\Delta_e^2}\right).
\end{equation}
The implied constant depends only on \(\alpha\) and is uniform in \(\Delta_e,\delta,\eta\).
\end{theorem}

Theorem~\ref{thm: lower bound} shows that the need for human expert evaluations is not merely a consequence of our calibration method. Even with unrestricted AI executions, adaptive evaluations, and more flexible deployment decisions, there are environments in which expert judgments about output quality cannot be replaced by additional AI observations. AI execution records can reveal how often answers pass verification, but they do not necessarily reveal whether those answers meet human experts' quality standards. Our result establishes that this distinction creates a fundamental requirement for human evaluation, regardless of how flexibly an organization operates its AI workflow.

The lower bound also establishes the minimax rate-optimality of our method in the key parameters governing human evaluation effort. Combining Theorems~\ref{thm: sufficient evaluations} and~\ref{thm: lower bound}, for fixed \(L,T,r_{\mathrm{tar}},\alpha\), and positive \(\Delta_r\), we obtain
\(
n^\star=\Theta\!\left(\frac{\log(1/\delta)+\log(1/\eta)}{\Delta_e^2}\right).
\)
Thus, our method achieves the optimal minimax rate in the quality margin \(\Delta_e\) and the two confidence levels, even against procedures allowing adaptive evaluation and general history-dependent deployment. This establishes that the inverse-square dependence on the quality margin reflects a fundamental difficulty of human quality assessment rather than inefficiency specific to our method. The rate matching does not extend to the number of thresholds, the target automation rate, or the automation-rate separation. In particular, the broader comparison class permits additional AI executions without human evaluations, allowing the lower bound to isolate the difficulty that fundamentally requires expert judgments.

\noindent\textbf{Proof sketch.}
Information-theoretic lower bounds typically construct hard environments that are difficult to distinguish and quantify the information needed to guarantee good performance across them \citep{polyanskiy2025information}. Our setting presents an additional challenge. Competing procedures may adaptively select tasks for evaluation, remember their quality labels, and use this information to make task-specific deployment decisions. Thus, a hard instance must remain difficult even when previously evaluated tasks can be treated differently during deployment.

To address this challenge, we construct two families of environments over a large population of \(M\) tasks. The generator always produces the same answer, and the verifier always assigns a passing score. The families differ only in human-assessed quality, with error rates \(p_0<\alpha-\Delta_e\) and \(p_1>\alpha\). Within each environment, the quality function is fixed, but incorrect answers are distributed uniformly across tasks. Let \(\mathbb Q_i\) denote the joint law induced by this random construction under family \(i\in\{0,1\}\) and the calibration procedure, and let \(D\) be the complete calibration transcript. After \(k\le n\) distinct tasks have been evaluated, with \(f\) errors, the conditional error probability of a previously unevaluated task is \(\zeta_i=(Mp_i-f)/(M-k)\), satisfying \(|\zeta_i-p_i|\le n/(M-n)\). Write \(\operatorname{KL}\) for the Kullback--Leibler divergence, where \(\operatorname{KL}(u,v)=u\log(u/v)+(1-u)\log((1-u)/(1-v))\) for Bernoulli parameters \(u,v\in(0,1)\). Since unlabeled AI observations are identical across the families, the KL chain rule gives
\[
\operatorname{KL}(\mathbb Q_i^D,\mathbb Q_{1-i}^D)
\le n\left(\operatorname{KL}(p_i,p_{1-i})+o_M(1)\right),
\]
where \(o_M(1)\to0\) as \(M\to\infty\) for fixed \(n,p_0,p_1\).

The construction also prevents selective deployment from avoiding this difficulty. Previously evaluated tasks account for at most \(n/M\) of the population, while the remaining error locations are exchangeable. Consequently, achieving a positive automation target requires releasing answers whose quality has not been observed. Let \(E\) denote the event that the selected rule achieves the automation target. In our construction, \(E\) is determined by the calibration transcript. Target attainment in the favorable family requires \(\mathbb Q_0(E)\ge1-\eta\), whereas quality protection in the unfavorable family implies \(\mathbb Q_1(E)\le\delta+o_M(1)\). Applying data processing in both directions, then letting \(M\to\infty\), \(p_0\uparrow\alpha-\Delta_e\), and \(p_1\downarrow\alpha\), yields
\[
\begin{aligned}
n &\ge
\max\left\{
\frac{\operatorname{KL}(1-\eta,\delta)}
{\operatorname{KL}(\alpha-\Delta_e,\alpha)},
\frac{\operatorname{KL}(1-\delta,\eta)}
{\operatorname{KL}(\alpha,\alpha-\Delta_e)}
\right\}\gtrsim
\frac{\log(1/\delta)+\log(1/\eta)}{\Delta_e^2}.
\end{aligned}
\]
The final bound follows by combining the two inequalities, with a constant depending only on \(\alpha\).% The Electronic Companion provides the complete proof.

%\clearpage
\section{Numerical Studies}
\label{sec: numerical}

In this section, we evaluate our method through a financial question-answering case study using the MultiHiertt dataset. We first examine deployment performance across a broad range of questions and then focus on a subset of particularly challenging tasks to study how quality requirements and expert evaluations affect safe automation. We also conduct synthetic experiments to assess whether the method behaves empirically as predicted by our theory.

\subsection{Case Study: Financial Question Answering}

We study financial question answering using MultiHiertt \citep{zhao2022multihiertt}, a benchmark based on financial reports containing text and multiple hierarchical tables. The questions range from retrieving financial figures and performing straightforward calculations to identifying relevant years or business segments and combining information across tables. This variation provides a setting for examining how much financial question answering an AI workflow can safely automate and which tasks remain with human experts.

MultiHiertt contains 10,440 questions, including 7,830 training, 1,044 development, and 1,566 test questions. We consider all 8,874 questions from the training and development sets. We also examine a subset of 1,676 challenging questions in Section~\ref{sec: challenging tasks} to study how our framework performs when AI faces particularly difficult tasks.

The following example illustrates a financial question that requires conditional information selection and calculation.

\begin{tcolorbox}[
    sharp corners,
    boxrule=0.5pt,
    colback=gray!5,
    fontupper=\small,
    before skip=10pt,
    after skip=10pt
]

\textbf{Illustrative Example:} ``What is the growing rate of Service of Mission Systems in the year
with the most Product of Mission Systems?''

\textbf{Source table excerpt:}
Sales for the Mission Systems segment, in millions of dollars.
Other segments and operating-cost columns are omitted from this excerpt.

\par\vspace{10pt}
\noindent\makebox[\linewidth][c]{%
\begin{tabular}{@{}lrrr@{}}
\toprule
Mission Systems & 2018 & 2017 & 2016 \\
\midrule
Product sales & 7,329 & 7,012 & 6,726 \\
Service sales & 4,380 & 4,458 & 4,435 \\
\bottomrule
\end{tabular}%
}

\par\vspace{10pt}

\noindent\textbf{Calculation:}
Product sales are highest in 2018. The corresponding year-over-year
growth in service sales is
\[
\frac{4{,}380-4{,}458}{4{,}458}\times100\%
\approx -1.75\%.
\]

\textbf{Reasoning required:}
Identify the Mission Systems segment, distinguish sales from operating
costs, select the year with the highest product sales, and calculate
growth using service sales in that year and the preceding year.

\end{tcolorbox}

\noindent\textbf{Data collection and evaluation.}
For each question, we run a ten-round generate--verify workflow. The generator, \texttt{gpt-5.4-mini} (reasoning effort \texttt{none}), produces an answer with supporting evidence or calculations. We set its temperature to zero to favor consistent responses over random variation, as desired in organizational applications. We use a stronger model, \texttt{gpt-6-luna} (reasoning effort \texttt{medium}), as the verifier to better identify errors and guide revisions. In subsequent rounds, the generator revises its response using only the preceding answer, support, score, and feedback, while the verifier evaluates each response without prior-round history. Neither model receives benchmark answers or annotated reasoning programs.

We collect complete ten-round trajectories for 8,872 of the 8,874 questions, regardless of verifier scores, yielding 88,720 answers and their corresponding verifier outputs. After excluding 377 questions whose answers could not be reliably determined through source-based review, the final dataset contains 8,495 questions and 84,950 answers. These trajectories allow us to evaluate different release thresholds and round limits using the same AI-generated responses. Data collection took approximately five hours in total, with an estimated API cost of \$317, including preliminary small-scale tests.

We assess answer correctness using the official MultiHiertt evaluator and an updated numerical grader. The original evaluator sometimes mishandles negative signs, units, and numerical precision, so we revise its grading rules accordingly. We further verify the grading results against the original financial reports and relevant calculations with Codex assistance. This review identifies both incorrectly rejected and incorrectly accepted answers. Ambiguous cases are examined by the researcher using the source evidence. The resulting corrections are applied consistently to repeated occurrences of the same question–answer pair. Each trajectory therefore contains ten verifier scores and corresponding error indicators, allowing us to evaluate the quality of answers released under different deployment policies.

\noindent\textbf{Comparison Method.}
Motivated by early-stopping methods that use consistency across repeated model outputs as a stopping signal \citep{li2024escape,mao2025early}, we consider \textbf{Verifier Consistency}, a heuristic that relies solely on verifier scores without expert evaluations. It releases the current answer when the verifier assigns scores of at least \(s\) for \(k\) consecutive rounds and otherwise escalates the task after ten rounds. We consider the same threshold set \(\mathcal S=\{0.90,0.95,1.00\}\) and \(k\in\{3,5\}\), yielding six policies evaluated using the same AI trajectories as our method.

\noindent\textbf{Preliminary Analysis.}
We first examine how repeated generation and verification affect answer correctness. Table~\ref{tab: round accuracy} reports the accuracy of the answers produced at each round, based on the human-defined quality labels for the 7,495 questions in the calibration pool.

\begin{table}[ht]
\centering
\caption{Answer accuracy across generation rounds.}
\label{tab: round accuracy}
\small
\setlength{\tabcolsep}{4pt}
\begin{tabular}{@{}l*{10}{r}@{}}
\toprule
Round $t$
& 1 & 2 & 3 & 4 & 5 & 6 & 7 & 8 & 9 & 10 \\
\midrule
Accuracy (\%)
& 53.60 & 78.31 & 83.39 & 85.70 & 86.87
& 87.54 & 87.99 & 88.26 & 88.53 & 88.65 \\
\bottomrule
\end{tabular}
\end{table}

Table~\ref{tab: round accuracy} shows that iterative revisions substantially improve answer correctness. Accuracy rises from 53.60\% in the first round to 78.31\% in the second round and reaches 88.65\% by the tenth round. Most of the improvement occurs in the earlier rounds, with smaller gains from subsequent revisions. Yet even after ten rounds, 11.35\% of the answers remain incorrect, exceeding the organization's 5\% error tolerance. Thus, while additional attempts improve AI performance, releasing every answer after ten rounds would not meet the required quality standard. Organizations must therefore determine which answers can be released without human review.

\noindent\textbf{Deployment Performance.}
We now examine how expert evaluations affect the automation and quality of deployed AI workflows. We randomly split the 8,495 questions into 7,495 for calibration and 1,000 for held-out evaluation. We set the quality tolerance to \(\alpha=0.05\), the confidence parameter to \(\delta=0.10\), and consider verifier thresholds \(\mathcal S=\{0.90,0.95,1.00\}\) with up to ten rounds. For each expert evaluation budget \(n\in\{1000,1500,2000,2500,3000,4000,5000\}\), we repeat calibration 30 times, sampling pilot questions without replacement. Each selected policy is evaluated on the same held-out questions. Table~\ref{tab: main deployment performance} compares our method with the verifier-based heuristics.

\begin{table}[t]
\centering
\caption{Held-out deployment performance of the proposed method and verifier-based heuristics.}
\label{tab: main deployment performance}
\small
\setlength{\tabcolsep}{5pt}
\begin{tabular}{@{}lrrrr@{}}
\toprule
Method & $n$ & Automation (SD) & Released error & 5\% requirement
\tabularnewline
\midrule
Proposed & 1,000 & 2.50\% (7.63) & 1.20\% & Meets
\tabularnewline
Proposed & 1,500 & 6.45\% (10.94) & 1.19\% & Meets
\tabularnewline
Proposed & 2,000 & 11.45\% (12.50) & 1.19\% & Meets
\tabularnewline
Proposed & 2,500 & 15.00\% (12.46) & 1.20\% & Meets
\tabularnewline
Proposed & 3,000 & 17.50\% (11.65) & 1.20\% & Meets
\tabularnewline
Proposed & 4,000 & 21.67\% (8.64) & 1.20\% & Meets
\tabularnewline
Proposed & 5,000 & 24.17\% (4.56) & 1.20\% & Meets
\tabularnewline
\midrule
Verifier Consistency $(0.90,3)$ & \textemdash & 86.90\% & \textbf{8.17\%} & \textbf{Violates}
\tabularnewline
Verifier Consistency $(0.90,5)$ & \textemdash & 81.60\% & \textbf{6.74\%} & \textbf{Violates}
\tabularnewline
Verifier Consistency $(0.95,3)$ & \textemdash & 82.10\% & \textbf{7.43\%} & \textbf{Violates}
\tabularnewline
Verifier Consistency $(0.95,5)$ & \textemdash & 76.90\% & \textbf{5.72\%} & \textbf{Violates}
\tabularnewline
Verifier Consistency $(1.00,3)$ & \textemdash & 1.30\% & 0.00\% & Meets
\tabularnewline
Verifier Consistency $(1.00,5)$ & \textemdash & 0.10\% & 0.00\% & Meets
\tabularnewline
\bottomrule
\end{tabular}
\par\smallskip
\parbox{0.94\linewidth}{\footnotesize
For the proposed method, automation rates are averaged over 30 calibration repetitions, with standard deviations in percentage points. Released error rates aggregate released answers across repetitions. The quality requirement is assessed on the held-out data. A dash indicates that a heuristic does not use a predeployment expert evaluation budget.
}
\end{table}

Table~\ref{tab: main deployment performance} first reveals the limitations of relying on repeated high verifier scores without human evaluation. At thresholds 0.90 and 0.95, Verifier Consistency automates between 76.90\% and 86.90\% of tasks, but all four policies violate the organization's 5\% error tolerance. Even requiring five consecutive scores of at least 0.95 results in an error rate of 5.72\%. Increasing the threshold to 1.00 satisfies the quality requirement, but automation falls sharply to 1.30\% with three consecutive scores and 0.10\% with five. These comparisons show that repeated high verifier scores do not necessarily establish the quality of released answers, while stricter consistency requirements can substantially limit automation.

Our method instead uses expert evaluations to determine which deployment policies meet the quality requirement. As shown in Table~\ref{tab: main deployment performance}, increasing the evaluation budget from 1,000 to 5,000 raises mean held-out automation from 2.50\% to 24.17\%. Throughout these budgets, released-output error rates remain around 1.20\%, well below the prescribed 5\% tolerance. Thus, additional expert evaluations allow the organization to automate more work without modifying the generator or verifier. Rather than relying on a fixed rule based solely on AI confidence, the organization can use human assessments to establish which deployment policies are sufficiently reliable. Note that the evaluation budgets in this study reflect the quality of the particular AI workflow, rather than a fixed requirement of our method. Improvements in the generator or verifier that reduce the error rate of released answers can increase the quality margin and, as our theory predicts, reduce the expert evaluations needed to support comparable automation.

The policies selected by our method also provide an interesting insight. With 1,000 expert evaluations, the method selects $(1.00,10)$ in only 3 of 30 repetitions, escalating all tasks in the remaining 27. At $n=5{,}000$, the same policy is selected in 29 of 30 repetitions. This policy allows the AI up to ten attempts to revise its answer, while requiring the verifier's maximum score before release. Its held-out automation rate is 25.00\%, with a released-output error rate of 1.20\%. Unlike Verifier Consistency, it does not require repeated high scores across consecutive rounds. Instead, our method uses human evaluations before deployment to establish whether a single verifier-accepted answer can be released safely.

Overall, these results highlight the importance of human evaluation in determining which AI-generated answers can be safely released. By calibrating deployment policies against expert assessments, our method enables organizations to make greater use of existing AI workflows while maintaining their quality requirements.

\subsubsection{Delegation on Challenging Tasks}
\label{sec: challenging tasks}

We next consider organizations seeking to automate particularly challenging tasks, where AI may struggle to produce reliable answers and human experts may need to handle much of the work. To study this setting, we focus on 1,676 challenging questions from MultiHiertt. These include 1,547 questions requiring at least three operations in their annotated reasoning programs and 129 two-operation questions involving conditional selection across different tables. Such questions require careful selection and calculation of financial quantities, where using the wrong year, item, or supporting evidence can produce an incorrect answer.

We first examine how the generator--verifier workflow performs on these questions. In the calibration pool of 1,476 questions, answer accuracy increases from 32.59\% in the first round to 78.46\% in the tenth round, showing that repeated revisions can improve correctness. However, additional rounds do not necessarily lead to reliable verifier-accepted answers. At threshold 0.90, 13.46\% of first-round passes are incorrect, compared with 24.68\% of answers first passing in later rounds. These results suggest that when AI struggles with difficult questions, repeated revisions may leave errors unresolved even when the verifier eventually assigns high confidence.

We consider three high verifier thresholds, 0.90, 0.95, and 1.00, with up to ten rounds. Table~\ref{tab: challenging policy quality} compares the automation and released-output error rates of representative policies on the calibration data. At thresholds 0.90 and 0.95, even first-round policies have error rates exceeding 10\%. At threshold 0.90, allowing a second round increases automation from 31.71\% to 62.87\%, but also raises the error rate from 13.46\% to 14.44\%. In contrast, requiring a score of 1.00 results in an error rate of only 1.48\% across ten rounds. Among the 238 answers that first reach this threshold after the first round, only two are incorrect. Thus, additional attempts can remain useful when the release requirement is sufficiently strict.

\begin{table}[ht]
\centering
\caption{Automation and released-output quality of representative policies on challenging financial questions.}
\label{tab: challenging policy quality}
\small
\begin{tabular}{@{}lrr@{}}
\toprule
Policy $(s,m)$ & Calibration automation & Calibration error
\tabularnewline
\midrule
$(0.90,1)$  & 31.71\% & 13.46\%
\tabularnewline
$(0.90,2)$  & 62.87\% & 14.44\%
\tabularnewline
$(0.95,1)$  & 29.40\% & 12.21\%
\tabularnewline
$(1.00,1)$  &  2.17\% &  6.25\%
\tabularnewline
$(1.00,10)$ & 18.29\% &  1.48\%
\tabularnewline
\bottomrule
\end{tabular}
\par\smallskip
\parbox{0.94\linewidth}{\footnotesize
Automation is the proportion of questions released by AI, and error is measured among released answers. Results are based on the 1,476 calibration questions.
}
\end{table}

We next apply our calibration method to examine how the organization's quality requirement affects automation in this challenging setting. We consider two error tolerances, $\alpha\in\{0.05,0.10\}$, while fixing $\delta=0.10$. For each tolerance, we consider evaluation budgets $n\in\{200,400,600,800\}$ and repeat calibration 30 times, sampling pilot questions without replacement. We use the same calibration repetitions and 200 held-out questions for both tolerances. Table~\ref{tab: challenging deployment} reports the results.

\begin{table}[t]
\centering
\caption{Expert evaluation budgets and deployment performance under two quality requirements.}
\label{tab: challenging deployment}
\small
\setlength{\tabcolsep}{5pt}
\begin{tabular}{@{}rrlrr@{}}
\toprule
$\alpha$
& $n$
& Selected policies (out of 30)
& Mean held-out automation (SD)
& Held-out error
\tabularnewline
\midrule
0.05 & 200 & $\pi^{\mathrm{esc}}$ 30
& 0.00\% (0.00) & ---
\tabularnewline
0.05 & 400 & $\pi^{\mathrm{esc}}$ 30
& 0.00\% (0.00) & ---
\tabularnewline
0.05 & 600 & $\pi^{\mathrm{esc}}$ 30
& 0.00\% (0.00) & ---
\tabularnewline
0.05 & 800 & $\pi^{\mathrm{esc}}$ 30
& 0.00\% (0.00) & ---
\tabularnewline
\midrule
0.10 & 200 & $\pi^{\mathrm{esc}}$ 30
& 0.00\% (0.00) & ---
\tabularnewline
0.10 & 400 & $\pi^{\mathrm{esc}}$ 30
& 0.00\% (0.00) & ---
\tabularnewline
0.10 & 600 & $\pi^{\mathrm{esc}}$ 16, $(1.00,10)$ 14
& 6.77\% (7.36) & 0.00\%
\tabularnewline
0.10 & 800 & $\pi^{\mathrm{esc}}$ 8, $(1.00,10)$ 22
& 10.63\% (6.52) & 0.00\%
\tabularnewline
\bottomrule
\end{tabular}
\par\smallskip
\parbox{0.94\linewidth}{\footnotesize
Standard deviations are in percentage points across 30 repetitions. Held-out error rates are reported for repetitions selecting a policy that releases answers. A dash indicates that no answers were released.
}
\end{table}

Table~\ref{tab: challenging deployment} shows that the quality requirement substantially affects the policies our method can approve with a given evaluation budget. Under the stricter tolerance $\alpha=0.05$, no candidate is approved in any of the 30 repetitions, even with 800 expert evaluations. When $\alpha=0.10$, the method begins selecting $(1.00,10)$ as the evaluation budget increases. This policy is selected in 14 of 30 repetitions at $n=600$ and 22 of 30 at $n=800$, raising mean held-out automation to 6.77\% and 10.63\%, respectively. No incorrect releases are observed among the selected policies.

The difference can be understood through the quality margin. As shown in Table~\ref{tab: challenging policy quality}, the calibration error rate of $(1.00,10)$ is 1.48\%, below both quality tolerances. However, its empirical quality margin is 3.52 percentage points when $\alpha=0.05$, compared with 8.52 percentage points when $\alpha=0.10$. The smaller margin makes it harder to establish that the policy meets the stricter requirement with the available expert evaluations. This agrees with our theoretical results, which show that the quality margin directly affects the expert effort needed for safe automation.

Overall, this experiment illustrates how an organization's quality requirement and available expert evaluations jointly affect AI delegation. Even when a deployment policy has a low observed error rate, the organization may lack sufficient evidence to approve it under a stringent quality standard. Our framework accounts for this distinction by escalating tasks to human experts when no policy can be approved. When the available evaluations support approval, the organization can automate more work while maintaining its chosen quality requirement.

\subsection{Synthetic Data Experiments}
\label{sec: synthetic studies}

We conduct synthetic experiments to complement the real-data case study by examining how AI workflow characteristics affect the expert evaluations needed for safe automation. Unlike in the real-data setting, synthetic experiments allow us to control automation rates and output quality separately. We first examine how the quality margin \(\Delta_e\) affects the required expert evaluations and then study the effect of automation-rate separation \(\Delta_r\).

\subsubsection{Effect of the Quality Margin on Expert Evaluations}
\label{sec: synthetic quality margin}

We first investigate how differences in output quality affect the expert evaluations needed to achieve the same target automation rate. We construct synthetic AI workflows with identical automation rates but different released-output error rates, allowing us to isolate the effect of the quality margin \(\Delta_e\).

We set the maximum number of rounds to \(T=10\) and consider two verifier thresholds, \(\mathcal S=\{0.80,0.95\}\). At each round, the verifier independently generates a score of 0.10, 0.85, or 0.99 with probabilities 0.50, 0.30, and 0.20, respectively. The corresponding error probabilities are 0.30, \(0.12-\Delta_e\), and \(0.07-\Delta_e\). We vary \(\Delta_e\) over \(\{0.01,0.02,0.03,0.04,0.05\}\), so the released-output error rate at threshold 0.80 is exactly \(0.10-\Delta_e\), while the automation rates remain unchanged. We set \(\alpha=\delta=0.10\) and target an automation rate of \(r_{\mathrm{tar}}=0.90\). The maximum automation rates at thresholds 0.80 and 0.95 are 99.90\% and 89.26\%, respectively, so only the former can achieve the target. For each setting, we run 1,000 Monte Carlo repetitions of our calibration method to evaluate how often the selected policy achieves the target automation rate.

\begin{figure}[ht]
\centering
\begin{minipage}{0.48\textwidth}
    \centering
    \includegraphics[width=\linewidth]{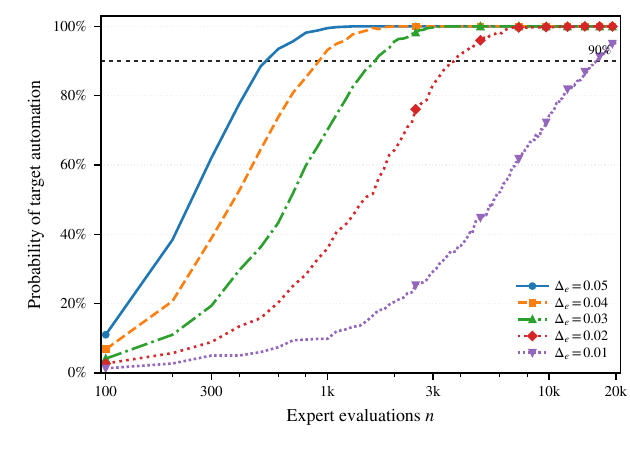}

    \small (a) Target automation probability
\end{minipage}
\hfill
\begin{minipage}{0.48\textwidth}
    \centering
    \includegraphics[width=\linewidth]{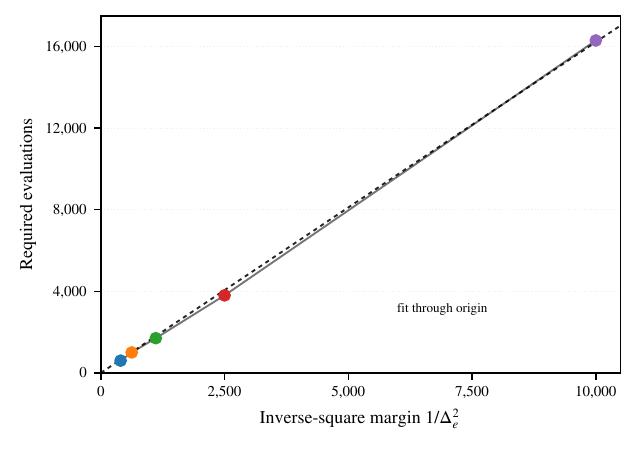}

    \small (b) Required expert evaluations
\end{minipage}
\caption{Effect of the quality margin on expert evaluation requirements. Panel (a) shows the probability of achieving the target automation rate as the evaluation budget increases. Panel (b) shows the minimum evaluation budget required to achieve the target with probability at least 90\%, plotted against \(1/\Delta_e^2\).}
\label{fig: synthetic quality margin}
\end{figure}

Figure~\ref{fig: synthetic quality margin}(a) shows that output quality substantially affects the ability to achieve the target automation rate with a given evaluation budget. For example, with \(n=2{,}000\) expert evaluations, the probability of achieving 90\% automation is 95.6\% when \(\Delta_e=0.03\), but only 20.5\% when \(\Delta_e=0.01\). Thus, even though the workflows have identical automation capabilities, the one with lower output quality requires considerably more expert evaluations.

Figure~\ref{fig: synthetic quality margin}(b) further examines this relationship by plotting the minimum evaluation budget needed to achieve the target automation rate with probability at least 90\% against \(1/\Delta_e^2\). The required budget increases from 600 evaluations at \(\Delta_e=0.05\) to 16,300 at \(\Delta_e=0.01\), an increase of more than 27 times. Across all five settings, the product of the required evaluation budget and \(\Delta_e^2\) remains approximately constant. This near-linear relationship agrees with the inverse-square dependence established by our theoretical results.

These findings reveal an important benefit of improving AI output quality. Even when the share of tasks that an AI workflow can automate remains unchanged, reducing its error rate can substantially reduce the human expert effort needed to support safe deployment. Thus, better output quality benefits organizations not only through more reliable answers but also through fewer expert evaluations required before deployment.

\subsubsection{Effect of Automation-Rate Separation on Expert Evaluations}
\label{sec: synthetic automation separation}

Even when all candidate policies meet the quality requirement, an organization may struggle to select a policy that achieves its desired automation rate. To examine this difficulty, we fix the released-output error rate of every candidate at 2\% and vary the automation-rate separation \(\Delta_r\) between a target-achieving policy and competing policies.

We set \(T=10\), \(\mathcal S=\{0.80,0.95\}\), and \(\alpha=\delta=0.10\), with \(r_{\mathrm{tar}}=0.90\). At each round, the verifier independently generates a score from \(\{0.10,0.85,0.99\}\), with probabilities \(0.10^{1/10}\), \((0.10+\Delta_r)^{1/10}-0.10^{1/10}\), and \(1-(0.10+\Delta_r)^{1/10}\), respectively. Each answer is incorrect with probability 0.02, independently of the verifier score. This construction gives \(r_{10}(0.80)=0.90\) and \(r_{10}(0.95)=0.90-\Delta_r\), while keeping the error rate of every candidate fixed at 2\%. We vary \(\Delta_r\) over \(\{0.01,0.02,0.03,0.04,0.05\}\) and run 1,000 Monte Carlo repetitions for each setting.

\begin{figure}[t]
\centering
\begin{minipage}{0.48\textwidth}
    \centering
    \includegraphics[width=\linewidth]{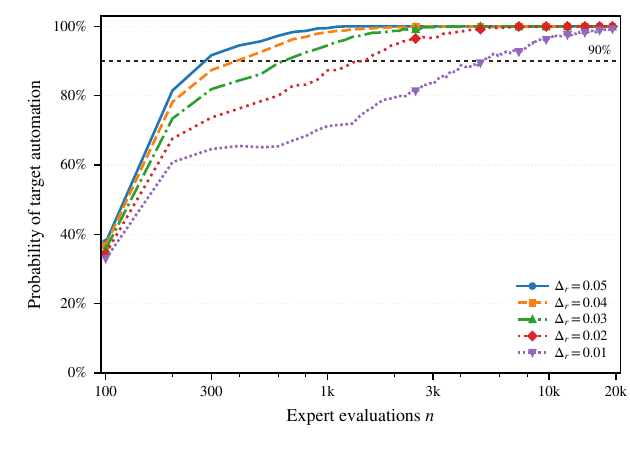}

    \small (a) Target automation probability
\end{minipage}
\hfill
\begin{minipage}{0.48\textwidth}
    \centering
    \includegraphics[width=\linewidth]{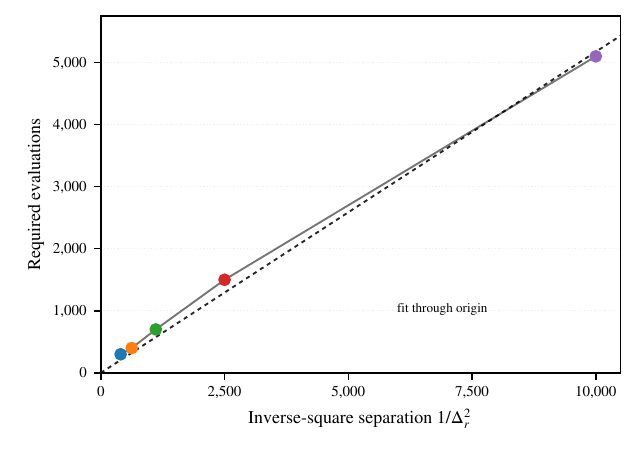}

    \small (b) Required expert evaluations
\end{minipage}
\caption{Effect of automation-rate separation on expert evaluation requirements. Panel (a) shows the probability of achieving the target automation rate as the evaluation budget increases. Panel (b) shows the minimum evaluation budget required to achieve the target with probability at least 90\%, plotted against \(1/\Delta_r^2\).}
\label{fig: synthetic automation separation}
\end{figure}

Figure~\ref{fig: synthetic automation separation}(a) shows that policies with similar automation rates are harder to distinguish, even though all candidates satisfy the quality requirement. For example, when \(\Delta_r=0.05\), the two full-horizon policies automate 90\% and 85\% of tasks, respectively. With \(n=1{,}000\) expert evaluations, our method achieves the target automation rate with probability 99.5\%. When the competing policy instead automates 89\% of tasks (\(\Delta_r=0.01\)), this probability falls to 71.2\%. Thus, even with identical output quality and the same target-achieving policy, a smaller difference in automation rates makes it harder to achieve the target.

Figure~\ref{fig: synthetic automation separation}(b) shows how this difficulty affects the required expert evaluations. The minimum evaluation budget needed to achieve the target with probability at least 90\% increases from 300 at \(\Delta_r=0.05\) to 5,100 at \(\Delta_r=0.01\), a 17-fold increase. Across the five settings, the required budget grows approximately with \(1/\Delta_r^2\), consistent with the dependence established by Theorem~\ref{thm: sufficient evaluations}.

This experiment shows that reliable AI outputs alone do not ensure that an organization can achieve its desired automation rate. The organization also needs enough AI execution records to distinguish deployment policies with similar automation rates. Although estimating these rates does not require human correctness labels, our calibration procedure collects execution records alongside expert evaluations. Increasing the evaluation budget therefore helps the organization identify policies that achieve its automation target. Together with the previous experiment, this illustrates the two statistical challenges in our framework: establishing that released answers meet the quality requirement and identifying a policy that provides sufficient automation.

\section{Conclusion and Discussion}
\label{sec: conclusion}

In this paper, we studied how organizations can safely delegate recurring tasks to an AI workflow. We develop a method that uses expert evaluations to calibrate black-box generate--verify workflows, selecting a verifier threshold and round limit without changing the generator or verifier. Our framework provides finite-sample quality guarantees and establishes sufficient evaluation requirements and a fundamental lower bound for achieving a target automation rate. In our financial question-answering case study, average held-out automation rises from 2.50\% to 24.17\% as expert evaluations increase, while released-output error rates remain around 1.20\%. In contrast, heuristics requiring consecutive high verifier scores either exceed the 5\% error tolerance or automate at most 1.30\% of tasks.

An important distinction is between what an AI workflow can achieve and what an organization has enough evidence to safely delegate to it. Low AI utilization may reflect either limited AI capability or insufficient evaluations to establish that a capable workflow meets the organization's quality standard. Improving the AI may allow it to handle more tasks and reduce the evaluations needed for safe deployment, while additional expert evaluations can help organizations better use its existing capabilities. Our theory also shows that workflows with similar automation capabilities can require substantially different numbers of expert evaluations depending on output quality. Organizations should therefore consider both AI performance and the human effort needed to validate its deployment. Investing expert effort before deployment can ultimately reduce the workload left to human experts afterward.

One direction for future research is to incorporate prior information about AI performance into calibration. AI systems are becoming increasingly specialized for particular industries and professional tasks, while evaluations of their performance across different applications are becoming more accessible \citep{burns2026vertical,grupen2026legal}. Organizations may therefore have access to benchmark results, previous deployments, or evaluations on related tasks before conducting their own pilot studies. Such information could help organizations allocate expert evaluations more efficiently and reduce the effort needed to validate AI deployment. An important direction is to develop methods that combine this prior information with new expert evaluations while preserving statistical guarantees under the organization's own quality standards.

Another direction is to extend our framework to broader AI governance requirements. As AI workflows become more complex and autonomous, organizations need to manage not only output quality and automation rates but also authorization, accountability, auditability, and human oversight \citep{wef2026agents,oecd2026agentic}. These considerations raise questions about which actions AI should be allowed to perform independently, when human approval is necessary, and how organizations can verify compliance with their governance standards. Extending our framework beyond release and error rates could provide a principled statistical approach to managing these additional requirements while balancing AI autonomy and human expert effort.

% --- References -------------------------------------------------------------
% For a clean compilation of the EMPTY template, this toggle is OFF.
% After pasting manuscript text and BibTeX records, set it to true.
\newif\ifIncludeBibliography
\IncludeBibliographytrue
\ifIncludeBibliography
  \bibliographystyle{abbrvnat}
  \bibliography{references}

@article{fugener2026roles,
  title={Roles of artificial intelligence in collaboration with humans: Automation, augmentation, and the future of work},
  author={F{\"u}gener, Andreas and Walzner, Dominik D and Gupta, Alok},
  journal={Management Science},
  volume={72},
  number={1},
  pages={538--557},
  year={2026},
  publisher={Informs}
}

@article{madaan2023self,
  title={Self-refine: Iterative refinement with self-feedback},
  author={Madaan, Aman and Tandon, Niket and Gupta, Prakhar and Hallinan, Skyler and Gao, Luyu and Wiegreffe, Sarah and Alon, Uri and Dziri, Nouha and Prabhumoye, Shrimai and Yang, Yiming and others},
  journal={Advances in neural information processing systems},
  volume={36},
  pages={46534--46594},
  year={2023}
}

@inproceedings{gou2024critic,
  title={Critic: Large language models can self-correct with tool-interactive critiquing},
  author={Gou, Zhibin and Shao, Zhihong and Gong, Yeyun and Yang, Yujiu and Duan, Nan and Chen, Weizhu and others},
  booktitle={International Conference on Learning Representations},
  volume={2024},
  pages={57734--57811},
  year={2024}
}

@article{angelopoulos2025learn,
  title={Learn then test: Calibrating predictive algorithms to achieve risk control},
  author={Angelopoulos, Anastasios N and Bates, Stephen and Cand{\`e}s, Emmanuel J and Jordan, Michael I and Lei, Lihua},
  journal={The Annals of Applied Statistics},
  volume={19},
  number={2},
  pages={1641--1662},
  year={2025},
  publisher={Institute of Mathematical Statistics}
}

@article{lee2024selective,
  title={Selective generation for controllable language models},
  author={Lee, Minjae and Kim, Kyungmin and Kim, Taesoo and Park, Sangdon},
  journal={Advances in Neural Information Processing Systems},
  volume={37},
  pages={50494--50527},
  year={2024}
}

@inproceedings{quach2024conformal,
  title={Conformal language modeling},
  author={Quach, Victor and Fisch, Adam and Schuster, Tal and Yala, Adam and Sohn, Jae Ho and Jaakkola, Tommi and Barzilay, Regina},
  booktitle={International Conference on Learning Representations},
  volume={2024},
  pages={11654--11681},
  year={2024}
}

@misc{anthropic2024agents,
  author = {{Anthropic}},
  title = {Building Effective Agents},
  year = {2024},
  month = dec,
  day = {19},
  howpublished = {\url{https://www.anthropic.com/engineering/building-effective-agents}},
  note = {Engineering blog post}
}

@book{polyanskiy2025information,
  title={Information theory: From coding to learning},
  author={Polyanskiy, Yury and Wu, Yihong},
  year={2025},
  publisher={Cambridge university press}
}

@incollection{boucheron2003concentration,
  title={Concentration inequalities},
  author={Boucheron, St{\'e}phane and Lugosi, G{\'a}bor and Bousquet, Olivier},
  booktitle={Summer school on machine learning},
  pages={208--240},
  year={2003},
  publisher={Springer}
}

@misc{mckinsey2025_agentic_ai_explained,
  author       = {{McKinsey \& Company}},
  title        = {Agentic AI Explained: When Machines Don't Just Chat, but Act},
  year         = {2025},
  month        = nov,
  day          = {25},
  howpublished = {\url{https://www.mckinsey.com/featured-insights/mckinsey-explainers/agentic-ai-explained-when-machines-dont-just-chat-but-act}},
  note         = {Article. Accessed 2026-05-10}
}

@article{holm1979simple,
  title={A simple sequentially rejective multiple test procedure},
  author={Holm, Sture},
  journal={Scandinavian journal of statistics},
  pages={65--70},
  year={1979},
  publisher={JSTOR}
}

@article{narasimhan2022post,
  title={Post-hoc estimators for learning to defer to an expert},
  author={Narasimhan, Harikrishna and Jitkrittum, Wittawat and Menon, Aditya K and Rawat, Ankit and Kumar, Sanjiv},
  journal={Advances in Neural Information Processing Systems},
  volume={35},
  pages={29292--29304},
  year={2022}
}

@article{snell2024scaling,
  title={Scaling llm test-time compute optimally can be more effective than scaling model parameters},
  author={Snell, Charlie and Lee, Jaehoon and Xu, Kelvin and Kumar, Aviral},
  journal={arXiv preprint arXiv:2408.03314},
  year={2024}
}

@article{kalayci2025optimal,
  title={Optimal Stopping vs Best-of-$ N $ for Inference Time Optimization},
  author={Kalayci, Yusuf and Raman, Vinod and Dughmi, Shaddin},
  journal={arXiv preprint arXiv:2510.01394},
  year={2025}
}

@inproceedings{jung2025trust,
  title={Trust or escalate: Llm judges with provable guarantees for human agreement},
  author={Jung, Jaehun and Brahman, Faeze and Choi, Yejin},
  booktitle={International Conference on Learning Representations},
  volume={2025},
  pages={3101--3125},
  year={2025}
}

@inproceedings{lai2022human,
  title={Human-ai collaboration via conditional delegation: A case study of content moderation},
  author={Lai, Vivian and Carton, Samuel and Bhatnagar, Rajat and Liao, Q Vera and Zhang, Yunfeng and Tan, Chenhao},
  booktitle={Proceedings of the 2022 CHI Conference on Human Factors in Computing Systems},
  pages={1--18},
  year={2022}
}

@article{raisch2021artificial,
  title={Artificial intelligence and management: The automation--augmentation paradox},
  author={Raisch, Sebastian and Krakowski, Sebastian},
  journal={Academy of management review},
  volume={46},
  number={1},
  pages={192--210},
  year={2021},
  publisher={Academy of Management Briarcliff Manor, NY}
}

@article{brynjolfsson2025generative,
  title={Generative AI at work},
  author={Brynjolfsson, Erik and Li, Danielle and Raymond, Lindsey},
  journal={The Quarterly Journal of Economics},
  volume={140},
  number={2},
  pages={889--942},
  year={2025},
  publisher={Oxford University Press}
}

@article{devigili2026skill,
  title={Skill Deprioritization: Reorganizing in the Age of Generative Artificial Intelligence},
  author={Devigili, Matteo and Yilmaz, Erdem Dogukan and Gaba, Vibha and Greve, Henrich},
  journal={Management Science},
  year={2026},
  publisher={INFORMS}
}

@article{fugener2022cognitive,
  title={Cognitive challenges in human--artificial intelligence collaboration: Investigating the path toward productive delegation},
  author={F{\"u}gener, Andreas and Grahl, J{\"o}rn and Gupta, Alok and Ketter, Wolfgang},
  journal={Information systems research},
  volume={33},
  number={2},
  pages={678--696},
  year={2022},
  publisher={INFORMS}
}

@article{lebovitz2022engage,
  title={To engage or not to engage with AI for critical judgments: How professionals deal with opacity when using AI for medical diagnosis},
  author={Lebovitz, Sarah and Lifshitz-Assaf, Hila and Levina, Natalia},
  journal={Organization science},
  volume={33},
  number={1},
  pages={126--148},
  year={2022},
  publisher={INFORMS}
}

@article{madras2018predict,
  title={Predict responsibly: improving fairness and accuracy by learning to defer},
  author={Madras, David and Pitassi, Toni and Zemel, Richard},
  journal={Advances in neural information processing systems},
  volume={31},
  year={2018}
}

@inproceedings{mozannar2020consistent,
  title={Consistent estimators for learning to defer to an expert},
  author={Mozannar, Hussein and Sontag, David},
  booktitle={International conference on machine learning},
  pages={7076--7087},
  year={2020},
  organization={PMLR}
}

@InProceedings{pmlr-v306-huang26bp,
  title = 	 {Delegation and Verification under {AI}},
  author =       {Huang, Lingxiao and Xiao, Wenyang and Vishnoi, Nisheeth K.},
  booktitle = 	 {Proceedings of the 43rd International Conference on Machine Learning},
  pages = 	 {48039--48071},
  year = 	 {2026},
  editor = 	 {Zhang, Tong and Dudik, Miroslav and Jaggi, Martin and Agarwal, Alekh and Li, Sharon and Schuurmans, Dale and Zhu, Jerry and Berkenkamp, Felix and Dong, Hanze and Bietti, Alberto},
  volume = 	 {306},
  series = 	 {Proceedings of Machine Learning Research},
  month = 	 {06--11 Jul},
  publisher =    {PMLR},
  url = 	 {https://proceedings.mlr.press/v306/huang26bp.html}
}

@article{srivastava2026governing,
  title={Governing AI-enabled decision making: Delegation, autonomy, and control},
  author={Srivastava, Abhishek},
  journal={PRODUCTION AND OPERATIONS MANAGEMENT},
  year={2026},
  publisher={SAGE PUBLICATIONS INC 2455 TELLER RD, THOUSAND OAKS, CA 91320 USA}
}

@article{dughmi2026adaptive,
  title={Adaptive Generate-Rank-Verify: Inference-Time Search with Costly Verification},
  author={Dughmi, Shaddin and Haghifam, Mahdi and Kalayci, Yusuf Hakan},
  journal={arXiv preprint arXiv:2605.17609},
  year={2026}
}

@InProceedings{pmlr-v306-wang26hd,
  title = 	 {Conformal Thinking: Risk Control for Reasoning on a Compute Budget},
  author =       {Wang, Xi and Suresh, Anushri and Zhang, Alvin and More, Rishi and Jurayj, William and Farajtabar, Mehrdad and Khashabi, Daniel and Nalisnick, Eric},
  booktitle = 	 {Proceedings of the 43rd International Conference on Machine Learning},
  pages = 	 {130032--130046},
  year = 	 {2026},
  editor = 	 {Zhang, Tong and Dudik, Miroslav and Jaggi, Martin and Agarwal, Alekh and Li, Sharon and Schuurmans, Dale and Zhu, Jerry and Berkenkamp, Felix and Dong, Hanze and Bietti, Alberto},
  volume = 	 {306},
  series = 	 {Proceedings of Machine Learning Research},
  month = 	 {06--11 Jul},
  publisher =    {PMLR},
  url = 	 {https://proceedings.mlr.press/v306/wang26hd.html}
}

@article{lee2026risk,
  title={Risk-Controlled Selective LLM Answering by Pricing Label-Free Checks},
  author={Lee, Dongyub Jude and Lee, Jungseob and Park, Chanjun and Moon, Hyeonseok and Lim, Heuiseok},
  journal={arXiv preprint arXiv:2609.37493},
  year={2026}
}

@techreport{agarwal2025designing,
  title={Designing human-ai collaboration: A sufficient-statistic approach},
  author={Agarwal, Nikhil and Moehring, Alex and Wolitzky, Alexander},
  year={2025},
  institution={National Bureau of Economic Research}
}

@article{giannitsas2026artificial,
  title={In artificial intelligence (AI) we (dis) trust? Navigating institutional pressures for automation and augmentation in the implementation of AI in organizations},
  author={Giannitsas, Dimitris and Sun, Ruonan and Baptista, Joao},
  journal={Information and Organization},
  volume={36},
  number={1},
  pages={100609},
  year={2026},
  publisher={Elsevier}
}

@article{zhong2026optimal,
  title={Optimal integration: Human, machine, and generative AI},
  author={Zhong, Hongda},
  journal={Management Science},
  volume={72},
  number={5},
  pages={3720--3739},
  year={2026},
  publisher={INFORMS}
}

@techreport{tinkoff2026state,
  title={The State of AI in 2026: On the Road to ROI},
  author={Tinkoff, Dan and Van der Veken, Lieven and Chui, Michael and Balakrishnan, Tara},
  institution={McKinsey \& Company},
  year={2026},
  month={aug},
  url={https://www.mckinsey.com/capabilities/quantumblack/our-insights/the-state-of-ai}
}

@techreport{langchain2026state,
  title={State of Agent Engineering},
  author={{LangChain}},
  institution={LangChain},
  year={2026},
  month={jun},
  url={https://www.langchain.com/state-of-agent-engineering}
}

@techreport{burns2026vertical,
  title={Gartner Standpoint: Vertical AI Revenue Will Significantly Outpace Horizontal AI},
  author={Burns, Ron},
  institution={Gartner},
  year={2026},
  month={may},
  url={https://www.gartner.com/en/documents/7890077}
}

@misc{grupen2026legal,
  title={Initial Results on Legal Agent Benchmark},
  author={Grupen, Niko and Pereyra, Gabe and Pereyra, Julio},
  howpublished={Harvey Research Blog},
  year={2026},
  month={may},
  url={https://www.harvey.ai/blog/legal-agent-benchmark-initial-results}
}

@techreport{wef2026agents,
  title={AI Agents in Action: A Playbook for Trusted Adoption, Authorization and Scaling},
  author={{World Economic Forum}},
  institution={World Economic Forum},
  year={2026},
  month={may},
  note={Developed in collaboration with Capgemini},
  url={https://www.weforum.org/publications/ai-agents-in-action-a-playbook-for-trusted-adoption-authorization-and-scaling/}
}

@techreport{oecd2026agentic,
  title={Agentic AI in Organisations: Early Insights from Practitioner Interviews},
  author={{OECD}},
  institution={OECD Publishing},
  series={OECD Artificial Intelligence Papers},
  number={65},
  year={2026},
  month={sep},
  doi={10.1787/1257a26f-en},
  url={https://doi.org/10.1787/1257a26f-en}
}

@article{noy2023experimental,
  title={Experimental evidence on the productivity effects of generative artificial intelligence},
  author={Noy, Shakked and Zhang, Whitney},
  journal={Science},
  volume={381},
  number={6654},
  pages={187--192},
  year={2023},
  publisher={American Association for the Advancement of Science}
}

@article{dell2026navigating,
  title={Navigating the jagged technological frontier: Field experimental evidence of the effects of artificial intelligence on knowledge worker productivity and quality},
  author={Dell’Acqua, Fabrizio and McFowland III, Edward and Mollick, Ethan and Lifshitz, Hila and Kellogg, Katherine C and Rajendran, Saran and Krayer, Lisa and Candelon, Fran{\c{c}}ois and Lakhani, Karim R},
  journal={Organization Science},
  volume={37},
  number={2},
  pages={403--423},
  year={2026},
  publisher={INFORMS}
}

@article{alves2024cost,
  title={Cost-sensitive learning to defer to multiple experts with workload constraints},
  author={Alves, Jean V and Leit{\~a}o, Diogo and Jesus, S{\'e}rgio and Sampaio, Marco OP and Li{\'e}bana, Javier and Saleiro, Pedro and Figueiredo, M{\'a}rio AT and Bizarro, Pedro},
  journal={arXiv preprint arXiv:2403.06906},
  year={2024}
}

@article{dell2026cybernetic,
  title={The cybernetic teammate: A field experiment on generative AI and teamwork},
  author={Dell’Acqua, Fabrizio and Ayoubi, Charles and Lifshitz, Hila and Sadun, Raffaella and Mollick, Ethan and Mollick, Lilach and Han, Yi and Goldman, Jeff and Nair, Hari and Taub, Stew and others},
  journal={Organization Science},
  volume={37},
  number={4},
  pages={1217--1242},
  year={2026},
  publisher={INFORMS}
}

@article{hegazy2026valid,
  title={Valid Stopping in Adaptive Generator-Verifier Loops},
  author={Hegazy, Mahmoud and Jordan, Michael I and Dieuleveut, Aymeric},
  journal={arXiv preprint arXiv:2610.06432},
  year={2026}
}

@inproceedings{zhao2022multihiertt,
  title={MultiHiertt: Numerical reasoning over multi hierarchical tabular and textual data},
  author={Zhao, Yilun and Li, Yunxiang and Li, Chenying and Zhang, Rui},
  booktitle={Proceedings of the 60th Annual Meeting of the Association for Computational Linguistics (Volume 1: Long Papers)},
  pages={6588--6600},
  year={2022}
}

@article{ahsen2026will,
  title={Will machines take over? Algorithms for human--machine collaborative decision making in healthcare},
  author={Ahsen, Mehmet E and Mookerjee, Radha and Ayvaci, Mehmet US},
  journal={Production and Operations Management},
  pages={10591478261443194},
  year={2026},
  publisher={SAGE Publications, Sage CA: Los Angeles, CA}
}

@article{ha2024artificial,
  title={Artificial Intelligence: Promise or Pitfalls? A Clinical Vignette of Real-Life ChatGPT Implementation in Perioperative Medicine: Ha and Kelley},
  author={Ha, Leslie Thienly and Kelley, Kristen D},
  journal={Journal of General Internal Medicine},
  volume={39},
  number={6},
  pages={1063--1067},
  year={2024},
  publisher={Springer}
}

@article{boucheron2013concentration,
  title={Concentration Inequalities: A Nonasymptotic Theory of Independence. Oxford Univ. Press, Oxford},
  author={Boucheron, S and Lugosi, G and Massart, P},
  year={2013}
}

@article{howard2020time,
  title={Time-uniform Chernoff bounds via nonnegative supermartingales},
  author={Howard, Steven R and Ramdas, Aaditya and McAuliffe, Jon and Sekhon, Jasjeet},
  year={2020}
}

@techreport{cfpb2023chatbots,
  title={Chatbots in Consumer Finance},
  author={{Consumer Financial Protection Bureau}},
  institution={Consumer Financial Protection Bureau},
  type={Issue Spotlight},
  year={2023},
  month={June},
  url={https://www.consumerfinance.gov/data-research/research-reports/chatbots-in-consumer-finance/chatbots-in-consumer-finance/}
}

@article{li2024escape,
  title={Escape sky-high cost: Early-stopping self-consistency for multi-step reasoning},
  author={Li, Yiwei and Yuan, Peiwen and Feng, Shaoxiong and Pan, Boyuan and Wang, Xinglin and Sun, Bin and Wang, Heda and Li, Kan},
  journal={arXiv preprint arXiv:2401.10480},
  year={2024}
}

@article{mao2025early,
  title={Early stopping chain-of-thoughts in large language models},
  author={Mao, Minjia and Yin, Bowen and Zhu, Yu and Fang, Xiao},
  journal={arXiv preprint arXiv:2509.14004},
  year={2025}
}

@article{ramdas2025hypothesis,
  title={Hypothesis testing with e-values},
  author={Ramdas, Aaditya and Wang, Ruodu},
  journal={Foundations and Trends{\textregistered} in Statistics},
  volume={1},
  number={1-2},
  pages={1--390},
  year={2025},
  publisher={Emerald Publishing Limited}
}

@article{li2021optimal,
  author  = {Li, Xiaoou and Chen, Yunxiao and Chen, Xi and Liu, Jingchen and Ying, Zhiliang},
  title   = {Optimal Stopping and Worker Selection in Crowdsourcing: An Adaptive Sequential Probability Ratio Test Framework},
  journal = {Statistica Sinica},
  year    = {2021},
  volume  = {31},
  number  = {1},
  pages   = {519--546},
  doi     = {10.5705/ss.202018.0300}
}

@article{lu2026contextual,
  author  = {Lu, Nan and Lee, Ethan and Fang, Ethan X. and Lu, Junwei},
  title   = {Contextual Online Uncertainty-Aware Preference Learning for Human Feedback},
  journal = {Journal of the American Statistical Association},
  year    = {2026},
  pages   = {1--25},
  doi     = {10.1080/01621459.2026.2668736},
  url     = {https://arxiv.org/abs/2504.19342}
}

@article{zhang2025self,
  author  = {Zhang, Shenao and Yu, Donghan and Sharma, Hiteshi and Zhong, Han and Liu, Zhihan and Yang, Ziyi and Wang, Shuohang and Awadalla, Hany Hassan and Wang, Zhaoran},
  title   = {Self-Exploring Language Models: Active Preference Elicitation for Online Alignment},
  journal = {Transactions on Machine Learning Research},
  year    = {2025},
  url     = {https://arxiv.org/abs/2405.19332}
}

@incollection{bhat2024value,
  title={Value alignment and trust in human-robot interaction: Insights from simulation and user study},
  author={Bhat, Shreyas and Lyons, Joseph B and Shi, Cong and Yang, X Jessie},
  booktitle={Discovering the frontiers of human-robot interaction: Insights and innovations in collaboration, communication, and control},
  pages={39--63},
  year={2024},
  publisher={Springer}
}

@misc{li2026coopa,
  title = {{COOPA}: A Modular {LLM} Agent Architecture for Operations Research Problems},
  author = {Li, Chuanhao and Xu, Xiaoan and Bergemann, Dirk and Fang, Ethan X. and Wei, Yehua and Yang, Zhuoran},
  year = {2026},
  eprint = {2606.27611},
  archivePrefix = {arXiv},
  url = {https://arxiv.org/abs/2606.27611}
}

@misc{zhang2025fisher,
  title = {Fisher Random Walk: Automatic Debiasing Contextual Preference Inference for Large Language Model Evaluation},
  author = {Zhang, Yichi and Belloni, Alexander and Fang, Ethan X. and Lu, Junwei and Xu, Xiaoan},
  year = {2025},
  eprint = {2509.05852},
  archivePrefix = {arXiv},
  url = {https://arxiv.org/abs/2509.05852}
}

@misc{liu2026reves,
  title = {{REVES}: {RE}vision and {VE}rification--Augmented Training for Test-Time Scaling},
  author = {Liu, Yuanxin and Zhou, Ruida and Zhao, Xinyan and Sharaf, Amr and Lin, Hongzhou and Biswas, Arijit and Ghavamzadeh, Mohammad and Wang, Zhaoran and Hong, Mingyi},
  year = {2026},
  eprint = {2606.18910},
  archivePrefix = {arXiv},
  url = {https://arxiv.org/abs/2606.18910}
}

@inproceedings{chen2025compo,
  title = {{ComPO}: Preference Alignment via Comparison Oracles},
  author = {Chen, Peter and Chen, Xi and Yin, Wotao and Lin, Tianyi},
  booktitle = {Advances in Neural Information Processing Systems},
  volume = {38},
  year = {2025},
  doi = {10.52202/085713-4065},
  url = {https://papers.nips.cc/paper_files/paper/2025/hash/b05bffeb1ef937677ef0e32f027b4c80-Abstract-Conference.html}
}

@article{chen2026stepwise,
  title = {Stepwise Guided Policy Optimization: Coloring Your Incorrect Reasoning in {GRPO}},
  author = {Chen, Peter and Li, Xiaopeng and Li, Ziniu and Chen, Xi and Lin, Tianyi},
  journal = {Transactions on Machine Learning Research},
  year = {2026},
  url = {https://www.jmlr.org/tmlr/papers/}
}

@inproceedings{xiao2025restoring,
  title = {Restoring Calibration for Aligned Large Language Models: A Calibration-Aware Fine-Tuning Approach},
  author = {Xiao, Jiancong and Hou, Bojian and Wang, Zhanliang and Jin, Ruochen and Long, Qi and Su, Weijie J. and Shen, Li},
  booktitle = {Proceedings of the 42nd International Conference on Machine Learning},
  series = {Proceedings of Machine Learning Research},
  volume = {267},
  pages = {68364--68390},
  publisher = {PMLR},
  year = {2025},
  url = {https://proceedings.mlr.press/v267/xiao25b.html}
}
\fi

% ============================================================================
% SUPPLEMENTARY MATERIAL: DP-conformal-inspired numbering and title.
% \BeginSupplementaryMaterial changes labels to Section A, Eq. (S1), etc.
% Keep this material after References, within the SAME compiled PDF.
% ============================================================================
\BeginSupplementaryMaterial

In this supplementary material, we provide detailed proofs and additional information supporting the results in the main text. Section~\ref{sec: proofs} provides complete proofs of the theoretical results in the main text.  Section~\ref{sec: numerical setup detail} describes the experimental setup, including real-data preprocessing, API configuration, and computational costs. 

\section{Proofs}\label{sec: proofs}
In this section, we provide the complete formal proofs for the theoretical results in the main text.

\subsection{Proof of Proposition~\ref{prop: round budgets}}
\label{sec: proof round budgets}

\begin{proof}
\noindent We first establish release ordering and then construct a single environment in which the released-output error rate is nonmonotone in both the round limit and the verifier threshold.

\noindent\textbf{Step 1. Establish release ordering.}
Fix an environment \(P\) and generate a full workflow trajectory \(H_T=(X,C_{1:T},S_{1:T})\) without stopping early. Since the generator and verifier follow the same conditional laws until stopping, truncating this trajectory at \(\tau_s\wedge m\) has the same distribution as execution under \(\pi^{(s,m)}\). Thus, \(r_m(s)=\mathbb P(\tau_s\le m)\). For \(s\le s'\) and \(m\le m'\), the first-pass events satisfy
\[
\begin{aligned}
\{\tau_{s'}\le m\}
&\subseteq\{\tau_s\le m\}
\subseteq\{\tau_s\le m'\},\\
r_m(s')&\le r_m(s)\le r_{m'}(s).
\end{aligned}
\]
This establishes release ordering without requiring independence across rounds.

\noindent\textbf{Step 2. Show nonmonotonicity in the round limit.}
Fix a threshold \(s\) and \(m<T\) with \(r_m(s)>0\). Let \(w=\mathbb P(\tau_s=m+1)\). If \(w=0\), the error rate remains unchanged. Otherwise, let \(\theta=\mathbb P(Q(X,C_{m+1})=0\mid\tau_s=m+1)\) be the error rate among newly released answers. Since earlier releases remain unchanged,
\[
\begin{aligned}
e_{m+1}(s)
&=\frac{r_m(s)e_m(s)+w\theta}{r_m(s)+w},\\
e_{m+1}(s)-e_m(s)
&=\frac{w}{r_{m+1}(s)}
\{\theta-e_m(s)\}.
\end{aligned}
\]
Thus, the error rate increases or decreases according to the quality of the additional answers relative to earlier releases.

To demonstrate nonmonotonicity, consider \(T=3\), \(\mathcal S=\{s_1,s_2,s_3\}=\{1/4,1/2,3/4\}\), and \(X\) uniformly distributed over three tasks \(x_1,x_2,x_3\). At round \(t\), the generator deterministically produces a distinct answer \(c_t\). For task \(x_i\), define the verifier score by \(S_t=(t/3)\mathbf1\{i\le t\}\) and human quality by \(Q(x_i,c_t)=\mathbf1\{t\ne2\}\). Thus, only the second-round answer fails the human quality standard.

At threshold \(s_1=1/4\), task \(x_i\) first passes at round \(i\). Consequently, \(r_m(s_1)=m/3\) and
\[
\bigl(e_1(s_1),e_2(s_1),e_3(s_1)\bigr)
=\left(0,\frac12,\frac13\right).
\]
The error rate first increases and then decreases as the round limit grows.

\noindent\textbf{Step 3. Show nonmonotonicity in the threshold.}
We use the same environment and fix \(m=3\). At threshold \(s_1\), tasks \(x_1,x_2,x_3\) release \(c_1,c_2,c_3\), respectively. At \(s_2\), tasks \(x_1,x_2\) release the incorrect answer \(c_2\), while \(x_3\) releases \(c_3\). At \(s_3\), every task releases the correct answer \(c_3\). Hence,
\[
\begin{aligned}
r_3(s_1)=r_3(s_2)=r_3(s_3)&=1,\\
\bigl(e_3(s_1),e_3(s_2),e_3(s_3)\bigr)
&=\left(\frac13,\frac23,0\right).
\end{aligned}
\]
Thus, the error rate is nonmonotone in the verifier threshold even when all three rules have the same automation rate. This completes the proof.
\end{proof}

\subsection{Proof of Theorem~\ref{thm: release quality}}
\label{sec: proof release quality}

\begin{proof}
\noindent Fix an environment \(P\) satisfying \(\sum_{j=1}^L r_T(s_j)>0\) and a positive integer \(n\). We suppress \(P\) in probabilities throughout the proof.

\noindent\textbf{Step 1. Establish the sampling law under random stopping.}
We show that the \(n\) evaluated records are i.i.d.\ from the distribution conditional on a verifier pass, despite the random number of pilot executions.

Construct an infinite sequence of independent pilot executions, each with an independent uniform threshold assignment. Let \(I_\ell=\mathbf1\{\tau_\ell\le T\}\) indicate whether execution \(\ell\) produces a verifier pass, and define \(Z_\ell=(J_\ell,\tau_\ell,Y_\ell)\) when \(I_\ell=1\), setting \(Z_\ell=\partial\) otherwise. The pairs \((I_\ell,Z_\ell)\) are i.i.d., with pass probability \(a=L^{-1}\sum_{j=1}^L r_T(s_j)>0\). Let \(\sigma_i\) be the execution index of the \(i\)th pass. The pilot stops at \(N=\sigma_n<\infty\) almost surely.

For arbitrary subsets \(D_1,\ldots,D_n\) of the evaluated-record space, let \(h_i\) count the failures preceding the \(i\)th pass. Independence across executions and summation over the waiting times give
\[
\begin{aligned}
&\mathbb P(Z_{\sigma_i}\in D_i,\ 1\le i\le n)\\
&=\sum_{h_1,\ldots,h_n\ge0}
\prod_{i=1}^n(1-a)^{h_i}\mathbb P(Z_1\in D_i)\\
&=\prod_{i=1}^n\frac{\mathbb P(Z_1\in D_i)}{a}\\
&=\prod_{i=1}^n\mathbb P(Z_1\in D_i\mid I_1=1).
\end{aligned}
\]
Hence, the \(n\) evaluated records are i.i.d.\ under the pass-conditional law. This argument accounts for random stopping without imposing independence across rounds within a task.

\noindent\textbf{Step 2. Derive the conditional binomial law.}
Fix a candidate \((s_j,m)\) with \(r_m(s_j)>0\), and write \(A=\sum_{h=1}^L r_T(s_h)\), \(v=r_m(s_j)/A\), and \(e=e_m(s_j)\). By Step 1, each evaluated record independently contributes an erroneous release, a correct release, or no release for this candidate, with probabilities \(ve\), \(v(1-e)\), and \(1-v\), respectively.

Consequently, the counts \((B_{j,m},R_{j,m}-B_{j,m},n-R_{j,m})\) follow a multinomial distribution. For \(0\le d\le k\le n\), its probability mass function factors as
\[
\begin{aligned}
&\mathbb P(R_{j,m}=k,B_{j,m}=d)\\
&=\binom nk\binom kd
(ve)^d\{v(1-e)\}^{k-d}(1-v)^{n-k}\\
&=\binom nk v^k(1-v)^{n-k}
\binom kd e^d(1-e)^{k-d}.
\end{aligned}
\]
It follows that \(R_{j,m}\sim\operatorname{Bin}(n,v)\) and \(B_{j,m}\mid R_{j,m}\sim\operatorname{Bin}(R_{j,m},e_m(s_j))\). If \(r_m(s_j)=0\), then \(R_{j,m}=0\) almost surely. In particular, the conditional binomial law requires no conditioning on the random execution count \(N\).

\noindent\textbf{Step 3. Verify the validity of the candidate p-values.}
Fix a candidate under the null hypothesis \(e=e_m(s_j)\ge\alpha\). Conditional on \(R_{j,m}=k>0\), Step 2 gives \(B_{j,m}\sim\operatorname{Bin}(k,e)\). Write \(F_{k,\theta}\) for the CDF of \(\operatorname{Bin}(k,\theta)\), so the p-value is \(p_{j,m}=F_{k,\alpha}(B_{j,m})\).

Since \(e\ge\alpha\), coupling Bernoulli variables through common uniform random variables gives \(F_{k,e}(d)\le F_{k,\alpha}(d)\) for every \(d\). For \(u\in[0,1]\), let \(d_u\) be the largest integer satisfying \(F_{k,\alpha}(d_u)\le u\), whenever such an integer exists. Then
\[
\begin{aligned}
&\mathbb P(p_{j,m}\le u\mid R_{j,m}=k)\\
&=\mathbb P(B_{j,m}\le d_u\mid R_{j,m}=k)\\
&=F_{k,e}(d_u)
\le F_{k,\alpha}(d_u)\le u.
\end{aligned}
\]
If no such \(d_u\) exists, the event is empty. When \(R_{j,m}=0\), the p-value is one. Averaging over \(R_{j,m}\) therefore gives \(\mathbb P(p_{j,m}\le u)\le u\) for every \(u\in[0,1]\). Thus, each true-null p-value is super-uniform under the full calibration law.

\noindent\textbf{Step 4. Establish simultaneous quality protection.}
The candidate p-values may be dependent because they share expert evaluations. We show that weighted Holm controls the probability of approving any true null, without requiring independence.

Let \(\mathcal I_0=\{(j,m):r_m(s_j)>0,\ e_m(s_j)\ge\alpha\}\) denote the true-null indices, and write \(W_0=\sum_{i\in\mathcal I_0}\omega_i\). Candidates with zero release probability receive p-value one and cannot be approved. If \(\mathcal I_0\) is empty, no false approval is possible.

Otherwise, let \(E\) denote the event that at least one true null is approved. On \(E\), consider the first true-null candidate \(i_k\) in the weighted Holm ordering. All true-null indices remain at positions \(k,\ldots,LT\), so the remaining weight satisfies \(W_k\ge W_0\). Since \(i_k\) is approved,
\[
p_{i_k}\le\frac{\delta\omega_{i_k}}{W_k}
\le\frac{\delta\omega_{i_k}}{W_0}.
\]
Consequently, using the super-uniformity established in Step 3,
\[
\begin{aligned}
\mathbb P(E)
&\le\mathbb P\left(
\bigcup_{i\in\mathcal I_0}
\left\{p_i\le\frac{\delta\omega_i}{W_0}\right\}
\right)\\
&\le\sum_{i\in\mathcal I_0}
\frac{\delta\omega_i}{W_0}
=\delta.
\end{aligned}
\]

On \(E^c\), every approved candidate has positive automation rate and error rate strictly below \(\alpha\). The selected rule therefore satisfies the population quality constraint whenever the approved set is nonempty. Otherwise, the algorithm selects the escalation-only rule, for which both the automation rate and bad-release probability are zero. Hence,
\[
\mathbb P_P\!\left(
b_P(\widehat\pi_n)\le\alpha r_P(\widehat\pi_n)
\right)\ge1-\mathbb P(E)\ge1-\delta.
\]
This completes the proof.
\end{proof}

\subsection{Proof of Theorem~\ref{thm: policy selection consistency}}
\label{sec: proof policy selection consistency}

\begin{proof}
\noindent Fix the environment and parameters in the theorem. Choose an optimal candidate \((s_{j_0},m_0)\) satisfying \(r_{m_0}(s_{j_0})=r_P^\star>0\) and \(e_{m_0}(s_{j_0})<\alpha\). Write \(A=\sum_{j=1}^L r_T(s_j)\) and \(a=A/L>0\). We suppress \(P\) in probabilities throughout the proof.

\noindent\textbf{Step 1. Establish consistency under random stopping.}
Construct an infinite sequence of independent pilot executions with independent uniform threshold assignments. Let \(I_\ell=\mathbf1\{\tau_\ell\le T\}\) indicate whether execution \(\ell\) produces an expert evaluation. For each positive integer \(n\), define \(N(n)=\inf\{q\ge1:\sum_{\ell=1}^q I_\ell=n\}\). Since the pass indicators are i.i.d.\ Bernoulli with mean \(a>0\), we have \(N(n)<\infty\) for every \(n\) and \(N(n)\to\infty\) almost surely.

Write \(N_j(n)\), \(R_{j,m}(n)\), and \(B_{j,m}(n)\) for the counts obtained by stopping at \(N(n)\). For each candidate \((s_j,m)\), the corresponding threshold-assignment, release, and erroneous-release indicators are bounded and i.i.d.\ across executions, with means \(1/L\), \(r_m(s_j)/L\), and \(b_m(s_j)/L\), respectively. The strong law of large numbers, applied along the random subsequence \(N(n)\), gives
\[
\frac{1}{N(n)}
\bigl(N_j(n),R_{j,m}(n),B_{j,m}(n)\bigr)
\xrightarrow{\mathrm{a.s.}}
\frac{1}{L}\bigl(1,r_m(s_j),b_m(s_j)\bigr).
\]
Since the candidate set is finite, these limits hold simultaneously. In particular, \(N_j(n)\to\infty\) and \(\widehat r_{m,n}(s_j)=R_{j,m}(n)/N_j(n)\to r_m(s_j)\) almost surely for every candidate. When \(r_m(s_j)>0\), we also have \(R_{j,m}(n)\to\infty\) and \(B_{j,m}(n)/R_{j,m}(n)\to e_m(s_j)\) almost surely.

\noindent\textbf{Step 2. Establish eventual approval and exclusion under weighted Holm.}
Fix a candidate with \(r_m(s_j)>0\), and write \(e=e_m(s_j)\), \(R_n=R_{j,m}(n)\), and \(B_n=B_{j,m}(n)\). Its exact binomial p-value is \(p_{j,m}(n)=\mathbb P_\alpha(K_{R_n}\le B_n)\), where \(K_{R_n}\sim\operatorname{Bin}(R_n,\alpha)\) under the auxiliary probability \(\mathbb P_\alpha\).

Suppose first that \(e<\alpha\), and set \(t=(e+\alpha)/2\). By Step~1, \(B_n/R_n<t\) for all sufficiently large \(n\), almost surely. Applying Hoeffding's inequality in Lemma~\ref{lem: hoeffding} gives
\[
\begin{aligned}
p_{j,m}(n)
&=\mathbb P_\alpha(K_{R_n}\le B_n)\\
&\le\mathbb P_\alpha(K_{R_n}\le R_nt)\\
&\le\exp\left(-\frac{R_n(\alpha-e)^2}{2}\right)
\longrightarrow0.
\end{aligned}
\]

If \(e>\alpha\), set \(t=(e+\alpha)/2\) again. Since \(B_n/R_n>t\) eventually, the upper-tail inequality in Lemma~\ref{lem: hoeffding} gives
\[
\begin{aligned}
1-p_{j,m}(n)
&=\mathbb P_\alpha(K_{R_n}>B_n)\\
&\le\mathbb P_\alpha(K_{R_n}\ge R_nt)\\
&\le\exp\left(-\frac{R_n(e-\alpha)^2}{2}\right)
\longrightarrow0.
\end{aligned}
\]
A candidate with \(r_m(s_j)=0\) has \(R_{j,m}(n)=0\) for every \(n\), almost surely, and therefore always receives p-value one.

We now translate these limits into decisions under weighted Holm. Every testing weight \(\omega_{j,m}\) is fixed and positive. For a strictly safe candidate, \(p_{j,m}(n)\to0\) implies \(p_{j,m}(n)\le\delta\omega_{j,m}\) eventually. This guarantees approval regardless of the other p-values. Indeed, if the candidate occupies position \(d\) in the weighted Holm ordering, then for every preceding position \(\ell\le d\),
\[
\frac{p_{i_\ell}}{\omega_{i_\ell}}
\le\frac{p_{i_d}}{\omega_{i_d}}
\le\delta
\le\frac{\delta}{W_\ell},
\]
where \(W_\ell\le1\) is the remaining testing weight. Thus, every candidate through position \(d\) satisfies its approval criterion.

Conversely, a candidate with \(e>\alpha\) eventually has \(p_{j,m}(n)>\delta\), while every weighted Holm cutoff is at most \(\delta\). Such candidates cannot be approved. The same holds for zero-release candidates. Since the candidate set is finite, almost surely, for all sufficiently large \(n\), the chosen strictly safe optimal candidate is approved and every approved candidate satisfies the population quality constraint. Candidates with \(e_m(s_j)=\alpha\) need not be excluded, since they remain population-feasible.

\noindent\textbf{Step 3. Establish optimal deployment recovery.}
By Step 1, the estimated automation rates converge uniformly over the finite candidate set. Define
\[
\varepsilon_n=
\max_{1\le j\le L,\;1\le m\le T}
\left|\widehat r_{m,n}(s_j)-r_m(s_j)\right|
\xrightarrow{\mathrm{a.s.}}0.
\]
If every feasible candidate attains \(r_P^\star\), Step 2 already establishes the result. Otherwise, the finite set of feasible suboptimal candidates has a strictly positive separation from the optimum. Let \(\gamma>0\) denote the smallest such difference.

On the probability-one event established in Steps 1 and 2, the chosen optimal candidate is eventually approved, every approved candidate is feasible, and \(\varepsilon_n<\gamma/3\). For any feasible candidate \((s_j,m)\) with \(r_m(s_j)<r_P^\star\),
\[
\begin{aligned}
&\widehat r_{m_0,n}(s_{j_0})
-\widehat r_{m,n}(s_j)\\
&\quad\ge
(r_P^\star-\varepsilon_n)
-(r_m(s_j)+\varepsilon_n)\\
&\quad\ge\gamma-2\varepsilon_n>0.
\end{aligned}
\]
Hence, empirical automation-rate maximization cannot select a feasible suboptimal candidate once \(n\) is sufficiently large.

Let \(E_n=\{b_P(\widehat\pi_n)\le\alpha r_P(\widehat\pi_n),\ r_P(\widehat\pi_n)=r_P^\star\}\). We have shown that \(E_n\) occurs eventually almost surely. Therefore, \(\mathbb P_P(E_n)\to1\). Since automation rates lie in \([0,1]\) and agree with \(r_P^\star\) on \(E_n\),
\[
\left|\mathbb E_P[r_P(\widehat\pi_n)]-r_P^\star\right|
\le\mathbb P_P(E_n^c)\longrightarrow0.
\]
Finally, if the optimal feasible candidate is unique, selecting a rule satisfying \(E_n\) is equivalent to selecting that candidate. Its selection probability therefore tends to one, completing the proof.
\end{proof}

\subsection{Proof of Theorem~\ref{thm: sufficient evaluations}}
\label{sec: proof sufficient evaluations}

\begin{proof}
\noindent Fix the parameters and an environment \(P\) satisfying the theorem. Choose a candidate \((s_j,m)\) satisfying its capability and separation conditions. Write \(r=r_m(s_j)\), \(e=e_m(s_j)\), and \(A=\sum_{h=1}^L r_T(s_h)\). Then \(r\ge r_{\mathrm{tar}}>0\), \(e\le\alpha-\Delta_e\), and \(r\le A\le L\). We suppress \(P\) in probabilities throughout the proof.

\noindent\textbf{Step 1. Ensure approval of a target-achieving candidate.}
We show that the chosen candidate receives enough expert evaluations to establish its quality and is approved by weighted Holm with high probability.

Write \(R=R_{j,m}\), \(B=B_{j,m}\), \(v=r/A\), and \(\mu=nv\). The conditional sampling law established in the proof of Theorem~\ref{thm: release quality} gives \(R\sim\operatorname{Bin}(n,v)\) and \(B\mid R\sim\operatorname{Bin}(R,e)\). Set \(t=\alpha-\Delta_e/2\), so \(t-e\ge\Delta_e/2\) and \(\alpha-t=\Delta_e/2\).

We first establish a binomial lower-tail bound for \(R\). Exponential Markov's inequality and the binomial generating function give
\[
\begin{aligned}
\mathbb P(R<\mu/2)
&\le 2^{\mu/2}\mathbb E[2^{-R}]\\
&=2^{\mu/2}(1-v/2)^n\\
&\le\exp\left(-\frac{1-\log2}{2}\mu\right)\\
&\le\exp(-\mu/8).
\end{aligned}
\]

Next, conditional on \(R=k>0\), applying Hoeffding's inequality in Lemma~\ref{lem: hoeffding} gives
\[
\begin{aligned}
\mathbb P(B>kt\mid R=k)
&\le\exp\left(-2k(t-e)^2\right)\\
&\le\exp(-k\Delta_e^2/2).
\end{aligned}
\]

Consider the event \(E_1=\{R\ge\mu/2,\ B\le tR\}\). Combining the preceding bounds yields
\[
\begin{aligned}
\mathbb P(E_1^c)
&\le\mathbb P(R<\mu/2)
+\mathbb P(R\ge\mu/2,\ B>tR)\\
&\le\exp(-\mu/8)
+\exp(-\mu\Delta_e^2/4).
\end{aligned}
\]

For each observed \(R>0\), let \(K_R\sim\operatorname{Bin}(R,\alpha)\) under the auxiliary probability \(\mathbb P_\alpha\). On \(E_1\), Hoeffding's inequality in Lemma~\ref{lem: hoeffding} gives
\[
\begin{aligned}
p_{j,m}
&=\mathbb P_\alpha(K_R\le B)\\
&\le\mathbb P_\alpha(K_R\le tR)\\
&\le\exp(-R\Delta_e^2/2)\\
&\le\exp(-\mu\Delta_e^2/4).
\end{aligned}
\]

Since every prescribed testing weight satisfies \(\omega_{h,k}\ge1/(3LT)\), the condition \(\mu\ge4\log(3LT/\delta)/\Delta_e^2\) ensures \(p_{j,m}\le\delta\omega_{j,m}\) on \(E_1\). By the weighted Holm approval argument in Step~2 of the proof of Theorem~\ref{thm: policy selection consistency}, this inequality guarantees approval regardless of the other p-values.

Consequently, under the stated condition on \(\mu\),
\[
\mathbb P\!\left(
(s_j,m)\notin\widehat{\mathcal C}_n
\right)
\le\exp(-\mu/8)+\exp(-\mu\Delta_e^2/4).
\]

\noindent\textbf{Step 2. Ensure sufficient executions under random stopping.}
Approval depends on expert evaluations, whereas comparing automation rates requires execution records. Since the pilot stops after \(n\) evaluations, the threshold-specific execution counts \(N_h\) are random. We establish a lower bound on all these counts using a deterministic prefix of the potential pilot sequence.

Let \(a=A/L\) be the probability that a pilot execution produces a verifier pass, and define \(t_0=\lceil n/(2a)\rceil\) and \(k_0=\lceil n/(4A)\rceil\). For threshold \(s_h\), let \(D_h(t_0)=\sum_{\ell=1}^{t_0}\mathbf1\{J_\ell=h\}\) count its assignments in the first \(t_0\) potential executions.

If \(N<t_0\), at least \(n\) passes must occur among the first \(t_0-1\) executions. Exponential Markov's inequality gives
\[
\begin{aligned}
\mathbb P(N<t_0)
&\le 2^{-n}(1+a)^{t_0-1}\\
&\le\exp\left(-n(\log2-1/2)\right)\\
&\le\exp(-n/6).
\end{aligned}
\]

Moreover, \(D_h(t_0)\sim\operatorname{Bin}(t_0,1/L)\), with mean \(t_0/L\). Applying the binomial lower-tail bound derived in Step~1 gives
\[
\begin{aligned}
\mathbb P\left(D_h(t_0)<\frac{t_0}{2L}\right)
&\le\exp\left(-\frac{t_0}{8L}\right)\\
&\le\exp\left(-\frac{n}{16A}\right).
\end{aligned}
\]

On \(N\ge t_0\), every assignment in the prefix has been executed. If also \(D_h(t_0)\ge t_0/(2L)\), then \(N_h\ge k_0\). Defining \(E_2=\{\min_{1\le h\le L}N_h\ge k_0\}\), a union bound yields
\[
\begin{aligned}
\mathbb P(E_2^c)
&\le\mathbb P(N<t_0)
+\sum_{h=1}^L
\mathbb P\left(D_h(t_0)<\frac{t_0}{2L}\right)\\
&\le\exp(-n/6)
+L\exp(-n/(16A)).
\end{aligned}
\]

This argument does not require independence between the stopping time and the threshold-assignment counts.

\noindent\textbf{Step 3. Control automation-rate estimates at random execution counts.}
The estimates \(R_{h,k}/N_h\) are evaluated at random sample sizes. We therefore establish a concentration bound that holds simultaneously over every sufficiently long prefix, rather than applying a fixed-sample bound directly at \(N_h\).

Consider i.i.d.\ Bernoulli variables \(Z_1,Z_2,\ldots\) with mean \(\vartheta\), and write \(\overline Z_q=q^{-1}\sum_{i=1}^q Z_i\). Let \(\mathcal F_q=\sigma(Z_1,\ldots,Z_q)\). Since \(Z_q-\vartheta\) has mean zero and lies in an interval of length one, Hoeffding's lemma (Lemma~\ref{lem: hoeffding lemma}) gives
\[
\mathbb E\left[e^{\lambda(Z_q-\vartheta)}\right]
\le\exp(\lambda^2/8)
\]
for every \(\lambda\in\mathbb R\).

For any \(\varepsilon>0\), define
\[
M_q^\pm
=
\exp\!\left(
\pm4\varepsilon\sum_{i=1}^q(Z_i-\vartheta)
-2q\varepsilon^2
\right),
\qquad M_0^\pm=1.
\]
By independence and Hoeffding's lemma,
\[
\begin{aligned}
\mathbb E[M_q^\pm\mid\mathcal F_{q-1}]
&=M_{q-1}^\pm e^{-2\varepsilon^2}
\mathbb E\left[
e^{\pm4\varepsilon(Z_q-\vartheta)}
\right]\\
&\le M_{q-1}^\pm
\exp\left(\frac{(4\varepsilon)^2}{8}
-2\varepsilon^2\right)\\
&=M_{q-1}^\pm.
\end{aligned}
\]
Thus, both processes are nonnegative supermartingales.

If \(\pm(\overline Z_q-\vartheta)\ge\varepsilon\) for some \(q\ge k_0\), then
\[
\begin{aligned}
\log M_q^\pm
&=\pm4q\varepsilon(\overline Z_q-\vartheta)
-2q\varepsilon^2\\
&\ge2q\varepsilon^2
\ge2k_0\varepsilon^2.
\end{aligned}
\]
Therefore, a deviation of magnitude at least \(\varepsilon\) forces one of the two supermartingales to cross \(\exp(2k_0\varepsilon^2)\). Applying Ville's inequality (Lemma~\ref{lem: ville inequality}) to both processes and taking a union bound gives
\[
\mathbb P\!\left(
\exists q\ge k_0:
|\overline Z_q-\vartheta|\ge\varepsilon
\right)
\le2\exp(-2k_0\varepsilon^2).
\]

To apply this bound, construct an independent sequence of fresh-task executions for each threshold, independent of the threshold-assignment sequence. Each assignment to \(s_h\) reveals the next unused execution in its sequence. This construction has the same law as the original pilot, including its random stopping time.

For a fixed candidate \((s_h,k)\), let \(\widehat r_{k,q}(s_h)\) be its empirical automation rate over the first \(q\) executions assigned to threshold \(s_h\). The corresponding release indicators are i.i.d.\ Bernoulli with mean \(r_k(s_h)\). Hence,
\[
\mathbb P\!\left(
\exists q\ge k_0:
|\widehat r_{k,q}(s_h)-r_k(s_h)|
\ge\varepsilon
\right)
\le2\exp(-2k_0\varepsilon^2).
\]

On \(E_2\), every realized \(N_h\) is at least \(k_0\), so this bound controls the actual estimates \(\widehat r_k(s_h)=\widehat r_{k,N_h}(s_h)\) without conditioning on the random execution counts.

\noindent\textbf{Step 4. Prevent selection of below-target candidates.}
Once the chosen candidate is approved, target attainment can fail only if empirical automation-rate maximization favors a below-target candidate. We use the separation condition to exclude these selections.

Let \(\mathcal H_-\) contain the thresholds \(h\ne j\) admitting at least one round limit with \(r_k(s_h)<r_{\mathrm{tar}}\). For each \(h\in\mathcal H_-\), define \(k_h^-=\max\{k:r_k(s_h)<r_{\mathrm{tar}}\}\). Release ordering implies that every below-target candidate at threshold \(s_h\) has both population and empirical automation rates no larger than those of \((s_h,k_h^-)\).

Suppose \(\mathcal H_-\ne\varnothing\). Let \(E_3\) be the event that the prefix estimates of the chosen candidate and all \((s_h,k_h^-)\), \(h\in\mathcal H_-\), deviate from their population rates by less than \(\Delta_r/3\) at every prefix length \(q\ge k_0\). There are at most \(L\) such candidates. By Step 3 and a union bound,
\[
\begin{aligned}
\mathbb P(E_3^c)
&\le2L\exp(-2k_0\Delta_r^2/9)\\
&\le2L\exp(-n\Delta_r^2/(18A)).
\end{aligned}
\]

On \(E_2\cap E_3\), every below-target candidate \((s_h,k)\) at another threshold satisfies
\[
\begin{aligned}
\widehat r_m(s_j)-\widehat r_k(s_h)
&\ge \widehat r_m(s_j)
-\widehat r_{k_h^-}(s_h)\\
&\ge r-r_{k_h^-}(s_h)
-\frac{2\Delta_r}{3}\\
&\ge\frac{\Delta_r}{3}>0.
\end{aligned}
\]
Thus, no below-target candidate at another threshold can defeat the chosen candidate.

At the chosen threshold \(s_j\), any below-target round limit \(k\) must satisfy \(k<m\), since \(r_m(s_j)\ge r_{\mathrm{tar}}\). Release ordering gives \(\widehat r_k(s_j)\le\widehat r_m(s_j)\), and our tie rule favors the larger round limit. Hence, these candidates also cannot be selected over the approved candidate.

Provided the approval condition from Step 1 holds, these comparisons give \(E_1\cap E_2\cap E_3\subseteq\{r_P(\widehat\pi_n)\ge r_{\mathrm{tar}}\}\). If \(\mathcal H_-=\varnothing\), no cross-threshold comparison is needed, and \(E_1\) alone ensures target attainment. In particular, this applies when \(L=1\).

\noindent\textbf{Step 5. Establish the sufficient evaluation budget.}
We combine the approval and comparison bounds to obtain an evaluation budget that guarantees target attainment with probability at least \(1-\eta\).

When \(\mathcal H_-\ne\varnothing\) and \(\mu\ge4\log(3LT/\delta)/\Delta_e^2\), Steps 1--4 give
\[
\begin{aligned}
&\mathbb P\!\left(
r_P(\widehat\pi_n)<r_{\mathrm{tar}}
\right)\\
&\le
\exp(-\mu/8)
+\exp(-\mu\Delta_e^2/4)\\
&\quad+\exp(-n/6)
+L\exp(-n/(16A))\\
&\quad+2L\exp(-n\Delta_r^2/(18A)).
\end{aligned}
\]

Set \(h_\delta=\log(3LT/\delta)\), \(h_\eta=\log(16/\eta)\), and \(h_L=\log(16L/\eta)\). A sufficient choice is
\[
n\ge128\left\{
\frac{A(h_\delta+h_\eta)}{r\Delta_e^2}
+\frac{Ah_L}{\Delta_r^2}
\right\}.
\]
Indeed, this condition implies \(\mu\ge4h_\delta/\Delta_e^2\). Substitution into the preceding failure bound gives \(\mathbb P(r_P(\widehat\pi_n)<r_{\mathrm{tar}})\le3\eta/8\le\eta\).

Since \(r\ge r_{\mathrm{tar}}\) and \(A\le L\), we may choose the uniform evaluation threshold
\[
\bar n=
\left\lceil128L\left\{
\frac{\log(48LT/(\delta\eta))}
{r_{\mathrm{tar}}\Delta_e^2}
+\frac{\log(16L/\eta)}
{\Delta_r^2}
\right\}\right\rceil.
\]
The restrictions \(\delta,\eta\le1/4\) ensure that the numerical constants inside the logarithms and the ceiling do not affect the claimed order. Therefore,
\[
\bar n
=O\!\left(
\frac{L\log(LT/(\delta\eta))}
{r_{\mathrm{tar}}\Delta_e^2}
+\frac{L\log(L/\eta)}
{\Delta_r^2}
\right).
\]

When \(\mathcal H_-=\varnothing\), the second term is unnecessary because approval alone guarantees target attainment. In particular, for \(L=1\), the separation condition can be removed and a sufficient threshold has order \(O(\log(T/(\delta\eta))/(r_{\mathrm{tar}}\Delta_e^2))\).

Finally, Theorem~\ref{thm: release quality} guarantees \(\mathbb P_P(b_P(\widehat\pi_n)\le\alpha r_P(\widehat\pi_n))\ge1-\delta\) for every fixed positive integer \(n\). Combining this with the target-attainment guarantee gives
\[
\mathbb P_P\!\left(
b_P(\widehat\pi_n)\le\alpha r_P(\widehat\pi_n),\
r_P(\widehat\pi_n)\ge r_{\mathrm{tar}}
\right)\ge1-\delta-\eta.
\]
This completes the proof.
\end{proof}

\subsection{Proof of Theorem~\ref{thm: lower bound}}
\label{sec: proof lower bound}

\begin{proof}
\noindent Fix the parameters in the theorem and let \(p=\alpha-\Delta_e\in[\alpha/2,\alpha)\). Consider any procedure satisfying the quality and target-automation guarantees over their respective environment classes, using at most \(n\) expert evaluations almost surely. We establish a lower bound on \(n\) that holds for every such procedure.

\noindent\textbf{Step 1. Construct hard environments with identical observable AI behavior.}
We construct two families that differ only in human-assessed quality, so additional AI executions cannot distinguish them.

Choose rational numbers \(0<p_0<p<\alpha<p_1<1\). Let \(M\) be a sufficiently large common multiple of their denominators, with \(M>n+1\), \(Mp_i>n\), and \(M(1-p_i)>n\) for \(i=0,1\). Select \(M\) distinct tasks from \(\mathcal X\), identify them with \([M]\), and let \(P_X\) be uniform on this set. The generator always returns a fixed answer \(c_0\), and the verifier always assigns score one.

For an error vector \(\theta\in\{0,1\}^M\), define the deterministic quality function \(Q_\theta(x,c)=1-\theta_x\) on the supported tasks, for every answer \(c\), and set \(Q_\theta=1\) elsewhere. Let \(\Theta_i=\{\theta:\sum_{x=1}^M\theta_x=Mp_i\}\), and let \(\Pi_i\) be uniform on \(\Theta_i\). Write \(\mathbb Q_i\) for the joint law obtained by drawing \(\theta\sim\Pi_i\) and running the procedure in the resulting environment \(P_\theta\).

Every candidate threshold--round-limit rule releases an answer in the first round. Thus, for every \(\theta\in\Theta_i\), we have \(r_m(s_j)=1\) and \(e_m(s_j)=p_i\). Every environment in the favorable family \(\Theta_0\) satisfies the capability condition of Theorem~\ref{thm: sufficient evaluations}, and its separation condition is vacuous because no candidate falls below the automation target. Both families have identical unlabeled AI observations.

\noindent\textbf{Step 2. Characterize the information available after adaptive evaluations.}
We show that adaptively choosing tasks does not reveal the locations of errors among previously unevaluated tasks.

Let \(U\) contain the procedure's randomization and potential unlabeled observations. Its distribution is identical under both families and independent of \(\theta\). At evaluation \(t\), let \(V_t\) denote the queried task and \(Y_t\) its error label. Repeated queries are permitted, and evaluations outside the support or after termination are represented by a dummy symbol \(\partial\). Define \(D_t=(U,V_1,Y_1,\ldots,V_t,Y_t)\) and \(D=D_n\).

Fix a feasible history \(D_t=d\). Let \(\mathcal O_t\) be the set of distinct evaluated tasks in \([M]\), with \(k=|\mathcal O_t|\), and let \(f\) be their number of errors. Conditional on \(d\), every vector in \(\Theta_i\) consistent with the observed labels has the same probability. Indeed, conditional on \(U\), the adaptive query decisions depend on \(\theta\) only through previously revealed labels. The remaining \(Mp_i-f\) errors are therefore uniformly distributed among the \(M-k\) unevaluated tasks.

Consequently, a newly evaluated supported task has conditional error probability
\[
\zeta_i(d)
=\frac{Mp_i-f}{M-k}
=p_i+\frac{kp_i-f}{M-k},
\qquad
|\zeta_i(d)-p_i|
\le\frac{n}{M-n}.
\]
This bound holds uniformly over adaptive histories. Repeated evaluations reveal previously observed labels and provide no new information about \(\theta\).

\noindent\textbf{Step 3. Show that selective deployment cannot avoid the difficulty.}
Even if the procedure remembers evaluated tasks and treats them differently during deployment, we show that it cannot reliably attain a positive automation target in the unfavorable family while maintaining quality protection.

For a complete calibration record \(d\), let \(a_x(d)\in[0,1]\) be the probability that the selected deployment rule releases an answer on task \(x\), integrating over deployment randomization. Since no new human-quality information is available during deployment, these release probabilities depend on \(d\) but not on the unobserved components of \(\theta\). Define \(r(d)=M^{-1}\sum_{x=1}^M a_x(d)\) and \(b_\theta(d)=M^{-1}\sum_{x=1}^M a_x(d)\theta_x\).

Let \(E=\{r(D)\ge r_{\mathrm{tar}}\}\) be the target-attainment event and \(G=\{b_\theta(D)\le\alpha r(D)\}\) the quality-protection event. The uniform deployment requirements imply \(\mathbb Q_0(E)\ge1-\eta\) and \(\mathbb Q_1(G^c)\le\delta\).

We now bound \(\mathbb Q_1(E)\). Fix \(D=d\), and write \(\mathcal O=\mathcal O_n(d)\), \(k=|\mathcal O|\), and \(f\) for the observed error count. Under \(\mathbb Q_1\), the unevaluated labels have conditional mean \(u=(Mp_1-f)/(M-k)\). Hence, writing \(\bar b(d)=\mathbb E_{\mathbb Q_1}[b_\theta(d)\mid D=d]\),
\[
\begin{aligned}
p_1r(d)-\bar b(d)
&=\frac1M\left\{
\sum_{x\in\mathcal O}a_x(d)(p_1-\theta_x)
+\frac{f-kp_1}{M-k}
\sum_{x\notin\mathcal O}a_x(d)
\right\}\\
&\le\frac{p_1(k-f)+(f-kp_1)_+}{M}\\
&=\frac{\max\{p_1(k-f),(1-p_1)f\}}{M}
\le\frac nM.
\end{aligned}
\]

The remaining labels are sampled uniformly without replacement. Thus, for distinct unevaluated tasks \(x,z\), their conditional covariance is \(-u(1-u)/(M-k-1)\le0\). Since the release weights are nonnegative and bounded by one,
\[
\begin{aligned}
\operatorname{Var}_{\mathbb Q_1}(b_\theta(d)\mid D=d)
&\le\frac{u(1-u)}{M^2}
\sum_{x\notin\mathcal O}a_x(d)^2\\
&\le\frac1{4M}.
\end{aligned}
\]

Let \(Z=b_\theta(D)-\bar b(D)\). Conditional centering and Cauchy--Schwarz give \(\mathbb E_{\mathbb Q_1}|Z|\le1/(2\sqrt M)\). On \(E\cap G\), the preceding mean bound implies \((p_1-\alpha)r_{\mathrm{tar}}\le n/M+|Z|\). Therefore,
\[
\begin{aligned}
\mathbb Q_1(E)
&\le\mathbb Q_1(G^c)+\mathbb Q_1(E\cap G)\\
&\le\delta+
\frac{n/M+\mathbb E_{\mathbb Q_1}|Z|}
{(p_1-\alpha)r_{\mathrm{tar}}}\\
&\le\delta+
\frac{n/M+1/(2\sqrt M)}
{(p_1-\alpha)r_{\mathrm{tar}}}
=:\delta+\varepsilon_M.
\end{aligned}
\]

For fixed \(n,p_1,r_{\mathrm{tar}}\), we have \(\varepsilon_M\to0\) as \(M\to\infty\). Thus, the target-attainment probabilities must differ substantially between the two families, even for procedures using task-specific deployment decisions.

\noindent\textbf{Step 4. Bound the information gained through adaptive evaluations.}
We now show that the complete calibration transcript contains at most \(n\) evaluations' worth of information about which family generated the environment.

Write \(\operatorname{KL}\) for the Kullback--Leibler divergence. For Bernoulli parameters \(u,v\in(0,1)\), define \(\operatorname{KL}(u,v)=u\log(u/v)+(1-u)\log((1-u)/(1-v))\).

Set \(\kappa=\frac12\min\{p_0,1-p_1\}>0\). By Step~2, for sufficiently large \(M\), the conditional error probabilities \(\zeta_0,\zeta_1\) belong to \([\kappa,1-\kappa]\), uniformly over all adaptive histories. The Bernoulli KL divergence has bounded first derivatives on this compact set. Consequently, for either ordered pair \(i\ne h\) in \(\{0,1\}\),
\[
\left|
\operatorname{KL}(\zeta_i,\zeta_h)
-\operatorname{KL}(p_i,p_h)
\right|
\le C_\kappa\frac{n}{M-n}
=:\xi_M,
\]
where \(C_\kappa\) depends only on \(\kappa\) and \(\xi_M\to0\) for fixed \(n,p_0,p_1\).

Conditional on \(D_{t-1}\), the next query is determined by the procedure. A newly evaluated supported task contributes \(\operatorname{KL}(\zeta_i,\zeta_h)\) to the conditional divergence, while repeated, outside-support, and padded queries contribute zero. Let \(K_t^{ih}(D_{t-1})\) denote this conditional divergence. Since the law of \(U\) is common to both families, applying the KL chain rule in Lemma~\ref{lem: kl chain rule} gives
\[
\begin{aligned}
\operatorname{KL}(\mathbb Q_i^D,\mathbb Q_h^D)
&=\sum_{t=1}^n
\mathbb E_{\mathbb Q_i}
[K_t^{ih}(D_{t-1})]\\
&\le n\{\operatorname{KL}(p_i,p_h)+\xi_M\}.
\end{aligned}
\]

It holds in both directions and includes all adaptive decisions and unlabeled observations.

\noindent\textbf{Step 5. Convert the information bound into an evaluation lower bound.}
The event \(E\) is determined by the calibration transcript \(D\), because the selected rule's automation rate depends only on its release behavior in this construction.

Write \(a_M=\mathbb Q_0(E)\) and \(b_M=\mathbb Q_1(E)\). Steps~3 and~4 give \(a_M\ge1-\eta\), \(b_M\le\delta+\varepsilon_M\), and finite transcript divergences in both directions. For sufficiently large \(M\), \(1-\eta>\delta+\varepsilon_M\). The finite divergences also imply mutual absolute continuity, so \(0<b_M<a_M<1\).

Applying the data processing inequality in Lemma~\ref{lem: kl data processing} to \(E\) under \(\mathbb Q_0,\mathbb Q_1\) gives
\[
\begin{aligned}
n\{\operatorname{KL}(p_0,p_1)+\xi_M\}
&\ge\operatorname{KL}(\mathbb Q_0^D,\mathbb Q_1^D)\\
&\ge\operatorname{KL}(a_M,b_M)\\
&\ge\operatorname{KL}(1-\eta,\delta+\varepsilon_M).
\end{aligned}
\]

Applying the same inequality to \(E^c\) in the reverse direction gives
\[
\begin{aligned}
n\{\operatorname{KL}(p_1,p_0)+\xi_M\}
&\ge\operatorname{KL}(\mathbb Q_1^D,\mathbb Q_0^D)\\
&\ge\operatorname{KL}(1-b_M,1-a_M)\\
&\ge\operatorname{KL}(1-\delta-\varepsilon_M,\eta).
\end{aligned}
\]
The final inequalities use the monotonicity of Bernoulli divergence when its first argument exceeds its second.

We first let \(M\to\infty\) along the admissible common multiples, keeping the procedure, \(n,p_0,p_1\), and deployment parameters fixed. Since \(\varepsilon_M,\xi_M\to0\), the two inequalities remain valid without the approximation terms. They hold for every admissible rational pair \(p_0,p_1\). Letting \(p_0\uparrow p=\alpha-\Delta_e\) and \(p_1\downarrow\alpha\) therefore gives
\[
n\ge
\max\left\{
\frac{\operatorname{KL}(1-\eta,\delta)}
{\operatorname{KL}(\alpha-\Delta_e,\alpha)},
\frac{\operatorname{KL}(1-\delta,\eta)}
{\operatorname{KL}(\alpha,\alpha-\Delta_e)}
\right\}.
\]
This bound applies to every procedure satisfying the two deployment requirements.

\noindent\textbf{Step 6. Derive the inverse-square evaluation rate.}
We convert the information lower bound into the claimed dependence on \(\Delta_e,\delta,\eta\).

For \(x,y\in(0,1)\), the inequality \(\log z\le z-1\) gives
\[
\begin{aligned}
\operatorname{KL}(x,y)
&\le x\left(\frac xy-1\right)
+(1-x)\left(\frac{1-x}{1-y}-1\right)\\
&=\frac{(x-y)^2}{y(1-y)}.
\end{aligned}
\]
Since \(p=\alpha-\Delta_e\ge\alpha/2\) and \(1-p\ge1-\alpha\), it follows that \(\operatorname{KL}(p,\alpha)\le\Delta_e^2/\{\alpha(1-\alpha)\}\) and \(\operatorname{KL}(\alpha,p)\le2\Delta_e^2/\{\alpha(1-\alpha)\}\).

For \(u,v\in(0,1/4]\), the binary entropy bound \(h(v)\le\log2\) similarly gives
\[
\begin{aligned}
\operatorname{KL}(1-v,u)
&=(1-v)\log\frac1u
+v\log\frac1{1-u}-h(v)\\
&\ge\frac34\log\frac1u-\log2
\ge\frac14\log\frac1u.
\end{aligned}
\]
Applying these inequalities to the two terms in Step 5 yields
\[
\begin{aligned}
n
&\ge\max\left\{
\frac{\alpha(1-\alpha)}{4\Delta_e^2}\log\frac1\delta,\,
\frac{\alpha(1-\alpha)}{8\Delta_e^2}\log\frac1\eta
\right\}\\
&\ge\frac{\alpha(1-\alpha)}{16\Delta_e^2}
\left\{\log\frac1\delta+\log\frac1\eta\right\}.
\end{aligned}
\]
Since the procedure was arbitrary, this proves the claimed minimax lower bound on \(n^\star\), with a constant depending only on \(\alpha\).

\noindent\textbf{Step 7. Match the lower bound with our method.}
The proposed algorithm belongs to the comparison class and satisfies both deployment requirements by Theorems~\ref{thm: release quality} and~\ref{thm: sufficient evaluations}. Thus, \(n^\star\le\bar n\), where the sufficient evaluation budget derived previously is
\[
\bar n=
\left\lceil128L\left\{
\frac{\log(48LT/(\delta\eta))}
{r_{\mathrm{tar}}\Delta_e^2}
+\frac{\log(16L/\eta)}
{\Delta_r^2}
\right\}\right\rceil.
\]

Let \(H=\log(1/\delta)+\log(1/\eta)\ge\log16\). For fixed \(L,T,r_{\mathrm{tar}},\alpha\), and positive \(\Delta_r\), the logarithmic terms are bounded by a constant multiple of \(H\). Since \(\Delta_e\le\alpha/2<1\), we obtain
\[
\begin{aligned}
\bar n
&\le1+C_0LH
\left\{
\frac1{r_{\mathrm{tar}}\Delta_e^2}
+\frac1{\Delta_r^2}
\right\}\\
&\le C\frac{H}{\Delta_e^2},
\end{aligned}
\]
where \(C_0\) and \(C\) depend only on the fixed parameters and are independent of \(\Delta_e,\delta,\eta\).

Combining this upper bound with Step 6 establishes
\[
n^\star
=\Theta\!\left(
\frac{\log(1/\delta)+\log(1/\eta)}{\Delta_e^2}
\right).
\]
Thus, our method attains the minimax evaluation rate in the quality margin and the two confidence levels, completing the proof.
\end{proof}

\subsection{Supporting Lemmas}\label{sec: supporting lemma}

\begin{lemma}[Full chain rule for KL divergence, Theorem~2.16(c) in \citealp{polyanskiy2025information}]
\label{lem: kl chain rule}
Let \(X_1,\ldots,X_n\) take values in standard Borel spaces, and let \(P\) and \(Q\) denote two probability laws for \(X_{1:n}=(X_1,\ldots,X_n)\). Then
\(
D(P_{X_{1:n}},Q_{X_{1:n}})
=
\sum_{i=1}^n
\mathbb E_{P}
\left[
D\left(
P_{X_i\mid X_{1:i-1}},
Q_{X_i\mid X_{1:i-1}}
\right)
\right].
\)
\end{lemma}

\begin{lemma}[Data processing inequality, Theorem~2.17 in \citealp{polyanskiy2025information}]
\label{lem: kl data processing}
Let \(P_Y=P_{Y\mid X}\circ P_X\) and \(Q_Y=P_{Y\mid X}\circ Q_X\), so that the conditional law \(P_{Y\mid X}\) is applied to \(P_X\) and \(Q_X\). Then
\[
D(P_Y,Q_Y)
\le
D(P_X,Q_X).
\]
If \(D(P_X,Q_X)<\infty\), equality holds if and only if
\[
D(P_{X\mid Y},Q_{X\mid Y}\mid P_Y)=0.
\]
\end{lemma}

\begin{lemma}[Hoeffding's inequality, Theorem~1 in \citealp{boucheron2003concentration}]
\label{lem: hoeffding}
Let \(X_1,\ldots,X_n\) be independent random variables such that \(X_i\in[a_i,b_i]\) almost surely, and let \(S_n=\sum_{i=1}^n X_i\). Then, for any \(t>0\),
\[
\mathbb P\left(
S_n-\mathbb E[S_n]\ge t
\right)
\le
\exp\left\{
-\frac{2t^2}{\sum_{i=1}^n(b_i-a_i)^2}
\right\},
\]
and
\[
\mathbb P\left(
S_n-\mathbb E[S_n]\le -t
\right)
\le
\exp\left\{
-\frac{2t^2}{\sum_{i=1}^n(b_i-a_i)^2}
\right\}.
\]
\end{lemma}

\begin{lemma}[Hoeffding's lemma, Lemma~2.2 in \citealp{boucheron2013concentration}]
\label{lem: hoeffding lemma}
Let \(X\) be a real-valued random variable satisfying \(\mathbb E[X]=0\) and \(a\le X\le b\) almost surely, where \(a<b\). Then, for every \(\lambda\in\mathbb R\),
\[
\mathbb E\left[e^{\lambda X}\right]
\le
\exp\left(\frac{\lambda^2(b-a)^2}{8}\right).
\]
\end{lemma}

\begin{lemma}[Ville's inequality, Lemma~1 in \citealp{howard2020time}]
\label{lem: ville inequality}
Let \((M_t)_{t\ge0}\) be a nonnegative supermartingale with respect to a filtration \((\mathcal F_t)_{t\ge0}\). Then, for every \(c>0\),
\[
\mathbb P\left(\sup_{t\ge0}M_t\ge c\right)
\le
\frac{\mathbb E[M_0]}{c}.
\]
\end{lemma}

\section{Additional Experimental Details}
\label{sec: numerical setup detail}

This section provides additional details on how we construct the generate--verify workflow and evaluate its outputs. We describe the prompts and information exchanged between the generator and verifier, followed by the grading and researcher adjudication procedures that determined  correctness.

\subsection{Workflow Configuration and Prompting}

We provide both the generator and verifier with the question and all accompanying financial-report text and tables from MultiHiertt, rather than only the annotated supporting evidence. Neither model receives benchmark answers, annotated reasoning programs, or correctness labels. The prompts are designed to produce answers that can be independently verified and revised based on specific feedback. Below, we present selected instructions from the production prompts, with the complete prompts available in the accompanying code.

\begin{tcolorbox}[
    sharp corners,
    boxrule=0.5pt,
    colback=gray!5,
    fontupper=\small,
    before skip=10pt,
    after skip=10pt
]
\textbf{Initial Generation Prompt}

%\medskip
\textit{Inputs:} Financial report and question.

%\medskip
\textit{Selected instructions:}

``When a question involves multiple conditions or tables, make sure the values used in your calculation correspond to the correct selected years or items.''

%\smallskip
``Use only the information in the provided financial report. Perform the necessary calculations carefully and check that the result addresses the exact question.''

%\smallskip
``The support should provide concrete, checkable information from the report, not an extended reasoning narrative.''

%\medskip
\textit{Outputs:} \texttt{answer} (final requested result) and \texttt{support} (brief supporting evidence and calculation).
\end{tcolorbox}

The supporting information identifies the financial quantities and calculations used to obtain the answer, allowing the verifier to check whether the generator selected the appropriate information rather than assessing the final answer alone.

\begin{tcolorbox}[
    breakable,
    sharp corners,
    boxrule=0.5pt,
    colback=gray!5,
    fontupper=\small,
    before skip=10pt,
    after skip=10pt
]
\textbf{Verification Prompt}

%\medskip
\textit{Inputs:} Financial report, question, current answer, and supporting information.

%\medskip
\textit{Selected instructions:}

``Treat the Generator's supporting explanation as a claim to be verified, not as independent evidence of correctness.''

%\smallskip
``An answer should not receive high confidence merely because it is consistent with some values in the report. Verify that the selected quantities, interpretation, and calculations correspond to what the question actually requires.''

%\smallskip
``Assign a confidence score between 0.00 and 1.00, rounded to two decimal places, representing your best assessment of whether the proposed final answer is correct.''

%\smallskip
``Do not reveal the correct final numerical answer, substitute the correct numerical values, or provide a complete worked-out solution. The feedback should guide the Generator to identify and correct its mistakes using the original report.''

%\medskip
\textit{Outputs:} \texttt{score} and \texttt{feedback} (at most 100 English words).
\end{tcolorbox}

The verifier is instructed to independently assess the answer against the financial report. High confidence requires positive verification rather than merely failing to identify an error. When errors or uncertainty remain, the feedback identifies the relevant issues without providing the correct answer directly.

\begin{tcolorbox}[
    breakable,
    sharp corners,
    boxrule=0.5pt,
    colback=gray!5,
    fontupper=\small,
    before skip=10pt,
    after skip=10pt
]
\textbf{Feedback-Guided Revision Prompt}

\medskip
\textit{Inputs:} Financial report, question, previous answer, previous support, previous verifier score, score interpretation, and feedback.

%\medskip
\textit{Selected instructions:}

``Review your previous answer and supporting evidence in light of the verifier's assessment.''

%\smallskip
``If you identify an error, correct the affected parts of your previous solution and recompute any results that depend on those corrections.''

%\smallskip
``Make sure your revision addresses the actual issue identified, rather than simply producing a different numerical answer.''

%\smallskip
``The verifier may also be mistaken. Do not change your answer solely because of a low score or negative feedback. If your previous answer remains supported by the financial report after careful examination, retain it.''

%\medskip
\textit{Outputs:} Updated \texttt{answer} and \texttt{support}.
\end{tcolorbox}

In each revision round, the generator receives only the immediately preceding answer, support, score, and feedback, together with a fixed verbal interpretation of the score. These interpretations distinguish high confidence (0.90--1.00), moderately high confidence (0.70--0.89), uncertainty (0.40--0.69), and low confidence (0.00--0.39). Earlier conversation history is not accumulated, and the verifier evaluates each current response without access to previous-round feedback.

\subsection{Answer Grading and Source-Based Review} 

MultiHiertt provides reference answers and an official evaluator. During our initial evaluation, we identified problems in numerical grading involving signs, units, and precision, as well as errors in some benchmark reference answers. We therefore updated the numerical grader and conducted additional source-based reviews to verify the grading results.

\noindent\textbf{Grading Criteria.} We evaluate the generator's final answer rather than its supporting explanation. The support may clarify its interpretation but cannot substitute for an incorrect answer. The updated grader preserves numerical signs and allows unit conversions supported by the financial context. Explicitly requested precision takes precedence. Otherwise, counts require exact values, monetary amounts are evaluated to the nearest whole requested unit, per-share amounts to two decimal places, percentages to one decimal place, and other computed quantities to two significant digits. Finer precision is also accepted when numerically correct.

\noindent\textbf{Source-Based Review and Adjudication.} We further verified the grading results using Codex with \texttt{gpt-6.1-sol} at high reasoning effort, checking answers against the original financial reports and relevant calculations. This review examined answers rejected by the updated grader and also identified incorrectly accepted answers. When the correctness of an answer remained ambiguous, the researcher examined the source evidence and resolved the case through additional review. Alternative interpretations were accepted only when supported by the source documents and consistent with the question's entities, periods, conditions, units, and required calculations. Decisions were applied consistently to repeated occurrences of the same question--answer pair.

\noindent\textbf{Exclusions.} Of the 8,874 questions in the training and development sets, two produced no responses because the API provider rejected their requests. We excluded another 377 questions whose answers could not be reliably determined after reviewing the source documents, including cases involving ambiguous or undefined calculations. For these questions, we excluded the entire ten-round trajectory rather than treating the answers as incorrect. The final dataset contains 8,495 questions and 84,950 answers. We retained the excluded records and documented the reasons for exclusion.

The following illustrate two problems encountered during correctness evaluation. The first concerns numerical parsing, while the second involves an error in the benchmark reference itself.

\medskip

\begin{tcolorbox}[
    breakable,
    sharp corners,
    boxrule=0.5pt,
    colback=gray!5,
    fontupper=\small,
    before skip=10pt,
    after skip=10pt
]
\textbf{Illustrative Example 1: Incorrect Handling of a Negative Answer}

\textbf{Question and source evidence:} The question requests the sum of ``Amortization of prior service cost (benefit)'' where recognized net actuarial loss is positive. Arconic's 2016 Annual Report provides qualifying pension amounts of 13, 16, and 18, together with other postretirement amounts of $-24$, $-37$, and $-25$ million.

\textbf{Calculation:}
\[
13+16+18-24-37-25=-39.
\]

\textbf{Grading issue and resolution:} The benchmark reference is also $-39$, but the original numerical parser removes the minus sign, converting the correct answer \texttt{-39} to $39$ before comparison. Preserving the sign resolves this error, and the answer of $-39$ million is accepted.
\end{tcolorbox}

%\clearpage
\begin{tcolorbox}[
    breakable,
    sharp corners,
    boxrule=0.5pt,
    colback=gray!5,
    fontupper=\small,
    before skip=10pt,
    after skip=10pt
]
\textbf{Illustrative Example 2: An Incorrect Benchmark Reference}

\textbf{Question and source evidence:} The question asks for the sum of qualified U.S.\ pension-plan components in the year with the largest U.S.\ Government and agency obligations. The relevant comparison selects 2008. Citigroup's 2010 Annual Report (p.~191) reports the pension-plan components for that year.

\textbf{Benchmark calculation:} The reference includes only benefits earned, interest cost, and expected asset return.
\[
23+674-949=-252.
\]

\textbf{Corrected calculation:}
Including the omitted prior service benefit of $-2$ and curtailment loss of $56$ gives
\[
23+674-949-2+56=-198.
\]

\textbf{Grading issue and resolution:} The corrected result agrees with the reported ``Net qualified (benefit) expense.'' After reviewing the source evidence, the researcher accepted $-198$ million as correct. Unlike the first example, this correction addresses an omitted component in the benchmark calculation rather than numerical formatting.
\end{tcolorbox}

\medskip

\noindent\textbf{Grading Results.} Source-based review and adjudication reclassified 45,275 answers initially marked incorrect as correct, while 190 answers initially marked correct were reclassified as incorrect. The resulting grading outcomes for the 84,950 retained answers are summarized below.

\par\vspace{8pt}
\noindent\makebox[\linewidth][c]{%
\begin{tabular}{@{}lrr@{}}
\toprule
Evaluation stage & Correct & Incorrect \tabularnewline
\midrule
Updated automatic grading & 25,117 & 59,833 \tabularnewline
After source review and adjudication & 70,202 & 14,748 \tabularnewline
\bottomrule
\end{tabular}%
}
\par\vspace{8pt}

\end{document}